\documentclass{article}

    \PassOptionsToPackage{numbers, compress}{natbib}
 \usepackage[main, final]{neurips_2026}
\usepackage[utf8]{inputenc} % allow utf-8 input
\usepackage[T1]{fontenc}    % use 8-bit T1 fonts
\usepackage{hyperref}       % hyperlinks
\usepackage{url}            % simple URL typesetting
\usepackage{booktabs}       % professional-quality tables
\usepackage{amsfonts}       % blackboard math symbols
\usepackage{nicefrac}       % compact symbols for 1/2, etc.
\usepackage{microtype}      % microtypography
\usepackage{xcolor}         % colors
\usepackage{lipsum}
\usepackage{caption}
\usepackage{graphicx}
\usepackage[table]{xcolor}
\usepackage{makecell}
\usepackage{amsmath}
\usepackage{tcolorbox}
\usepackage{multirow}
\usepackage{enumitem}
\usepackage{pifont}
\usepackage{duckuments}
\usepackage{wrapfig}
\usepackage{nicefrac}
\definecolor{codegreen}{rgb}{0,0.6,0}
\definecolor{codegray}{rgb}{0.7,0.7,0.7}
\definecolor{codepurple}{rgb}{0.58,0,0.82}
\definecolor{backcolour}{rgb}{1.0,1.0,1.0}
\definecolor{Light}{rgb}{0.21,0.49,0.74}
\definecolor{upcolor}{RGB}{57,182,74}

\definecolor{g_sum}{HTML}{3479E5}
\newcommand{\model}{Imagine3D-LLM}

\title{Imagine3D-LLM: Teaching MLLMs to \\ Imagine 3D Scenes Before Answering}

\author{
    Jaewoo Jung\textsuperscript{\rm 1,2}\thanks{Work done during a visiting researcher period at ETH Z\"urich} \quad
    Hyeonseo Yu\textsuperscript{\rm 1} \quad
    Honggyu An\textsuperscript{\rm 1} \quad
    Jisang Han\textsuperscript{\rm 1} \quad
    Mungyeom Kim\textsuperscript{\rm 1} \quad \\
    {\bf Minkyeong Jeon\textsuperscript{\rm 1} \quad
    Heeseong Shin\textsuperscript{\rm 1} \quad
    WonJun Moon\textsuperscript{\rm 1} \quad 
    Federico Tombari\textsuperscript{\rm 3,4} \quad} \\ 
    {\bf Daniel Barath\textsuperscript{\rm 2} \quad
    Marc Pollefeys\textsuperscript{\rm 2}$^{\dagger}$ \quad
    Seungryong Kim\textsuperscript{\rm 1}$^{\dagger}$ \quad
    Sunghwan Hong\textsuperscript{\rm 2,5}$^{\dagger}$} \\[5pt]
    \textsuperscript{\rm 1}KAIST AI \qquad \textsuperscript{\rm 2}ETH Z\"urich \qquad \textsuperscript{\rm 3}Google \qquad \textsuperscript{\rm 4}TUM \qquad \textsuperscript{\rm 5}ETH AI Center\\[3pt]
{\tt \href{https://cvlab-kaist.github.io/Imagine3D-LLM}%
    {\url{https://cvlab-kaist.github.io/Imagine3D-LLM}}}
}

\begin{document}

\maketitle

\begin{abstract}
  Reasoning about the 3D world from multi-view images remains a fundamental challenge for Multimodal Large Language Models (MLLMs). While modern MLLMs handle single-image inputs effectively, they struggle to integrate evidence across viewpoints into a coherent 3D understanding. A growing body of work attempts to close this gap by injecting 3D awareness into MLLMs, either by boosting fine-grained pixel-level cross-view correspondence or by fusing features from 3D geometry foundation models, yet a substantial gap to human reasoning persists. In this work, we revisit human spatial reasoning, which suggests that rather than relying on fine-grained geometry cues, humans roughly identify common objects across views, infer the relative geometry between viewpoints, and assemble a coarse 3D layout of the scene. Inspired by this process, we introduce \textbf{Imagine3D-LLM}, an MLLM that learns to assemble a similar compact 3D representation of the scene and conditions its answer on this representation. Concretely, we append a small set of learnable summary tokens after the image tokens, decode them into a compact 3D Gaussian Splatting representation supervised by a photometric reconstruction loss, and train jointly with the standard next-token prediction objective. Notably, although only the summary tokens receive direct reconstruction supervision, this objective also induces stronger cross-frame correspondence within the LLM's underlying image features, suggesting that learning to reconstruct propagates 3D-aware signals throughout the model. As a result, Imagine3D-LLM consistently outperforms prior approaches across multiple spatial reasoning and 3D understanding benchmarks, suggesting that imagining the scene can be more effective than being told its pixel-wise geometry.
\end{abstract}

\section{Introduction}
Reasoning about 3D structure from multi-view images is a fundamental problem in computer vision, underpinning a wide range of downstream applications such as robotics~\cite{shen2023distilled}, embodied AI~\cite{zhao2023learning}, AR/VR~\cite{brachmann2023accelerated}, and scene understanding~\cite{an2025c3g}. Driven by rapid progress in Multimodal Large Language Models (MLLMs)~\cite{yang2025qwen3,qwen2_5_vl, gpt4o, gemini_pro}, machines can now interpret a single image or a continuous video stream at a level that often rivals, and in some cases surpasses, human performance~\cite{yue2024mmmu, lu2024mathvista}. However, when the input shifts from a single viewpoint to a set of multi-view images that demand genuine 3D reasoning, even the largest and most capable models fall short of human-level competence~\cite{vsibench}.

A growing body of work has attempted to close this gap by injecting strong 3D priors into MLLMs. One line of research implicitly or explicitly boosts pixel-level correspondences across views by either embedding scene point clouds' 3D coordinates into patch-level features~\cite{video3dllm, llava3d,wang2025ross3d}, augmenting images with explicit visual markers~\cite{gpt4scene}, distilling features with accurate correspondences~\cite{huang20253drs}, or designing pretext tasks that encourage the model to align overlapping content across frames~\cite{wang2025ross3d}. A complementary line~\cite{zheng2025learning,fan2025vlm,hu2025g2vlm} fuses MLLM image features with ones from recently developed 3D reconstruction foundation models such as CUT3R~\cite{wang2025continuous} and VGGT~\cite{wang2025vggt}. While both approaches attempt to improve 3D awareness of MLLMs with {fine-grained pixel-level} 3D signals, these approaches yield only incremental gains~\cite{vsibench,spar}.

In this work, we take a step back and ask whether such \emph{fine-grained} 3D signals are truly the most direct route to improved 3D reasoning. To answer this, we revisit how humans actually perceive 3D structure from multi-view observations. Cognitive studies suggest that humans do not maintain pixel-perfect correspondences or dense depth maps in their heads~\cite{burgess2006spatial,shepard1971mental}. When presented with several views of a scene, humans instead identify common objects across views, use them as anchors to infer \emph{coarse and abstract} spatial relationships between viewpoints, and mentally assemble a compact layout of the scene in 3D. Therefore, we hypothesize that the representation that supports their downstream reasoning is rather object-level and approximate, not pixel-accurate.

Inspired by this human thinking process, we present \textbf{\model}, a framework that enables an MLLM to build an analogous \emph{compact 3D reconstruction} of the scene before answering. To enable this, we introduce a \emph{compact} set of learnable \textbf{Gaussian summary tokens}, which are appended to the sequence of image tokens fed into the LLM. From these tokens, we decode the parameters of 3D Gaussian Splatting (3DGS)~\cite{kerbl20233d} primitives, so that each Gaussian summary token corresponds to a small set of 3D Gaussians that together explain a portion of the scene.

To efficiently and effectively enable MLLMs to build an abstract and compact reconstruction of the scene, our Gaussian estimation pipeline incorporates two deliberate design choices. \textbf{(1)} To prevent the model from simply copy-pasting 2D image content into image-aligned Gaussians rather than recovering the underlying geometry, we predict the Gaussians from the dedicated Gaussian summary tokens instead of predicting them directly from the image features. \textbf{(2)} We impose an information bottleneck by using fewer Gaussian summary tokens than image tokens, which compels overlapping content across views to be merged into a shared set of tokens, encouraging the model to reason about which objects recur across views and how they fit together within a unified 3D layout. 

We jointly train \model\ with a photometric reconstruction loss on the rendered Gaussians and the standard next-token prediction loss for language modeling. Surprisingly, although only the Gaussian summary tokens receive direct reconstruction supervision, our analysis shows that the LLM's image features themselves become more 3D-aware: corresponding image features show more sharp attentions, where PCA visualizations further reveal that image features become more semantically organized, with the same object encoded consistently across views. These analyses validate that requiring the model to infer the underlying structure of the scene propagates 3D-aware signals throughout the LLM, reshaping its internal representations toward a coherent, object-centric 3D understanding. As a result, \model\ achieves significant performance gains across seven diverse spatial reasoning and 3D scene understanding benchmarks, suggesting that abstract scene reconstruction is a powerful inductive bias for building 3D-aware MLLMs, and that imagining the scene before answering can be more effective than being told its pixel-wise geometry.
\section{Related Works}

\paragraph{3D Scene Understanding.}
Grounding natural language in 3D environments is a long-standing goal of the 3D scene understanding community, supporting tasks such as 3D visual grounding~\cite{scanrefer, zhang2023multi3drefer}, 3D dense captioning~\cite{scan2cap, chen2023end, chen2024vote2cap}, and 3D question answering~\cite{scanqa, ma2022sqa3d, ma2026real3dqa, vsibench}. A prominent line of work distills features from 2D foundation models (most commonly CLIP~\cite{radford2021learning} or LSeg~\cite{li2022language}) into a reconstructed scene representation, including point clouds~\cite{yang2024regionplc, zhou2025ov3d, conceptfusion}, NeRFs~\cite{lerf2023, engelmann2024opennerf}, and more recently 3D Gaussian Splatting~\cite{jun2025dr, qin2024langsplat, shi2024language}, enabling open-vocabulary querying by matching distilled embeddings to a text prompt. These pipelines, however, typically rely on task-specific architectures and per-scene optimization, and their language understanding is bounded by the compositional capacity of the underlying 2D embedding model. We instead extend the language and visual reasoning of pretrained MLLMs to multi-view inputs, allowing a single generalist model to address these tasks through natural language without per-task or per-scene optimization.\vspace{-5pt}

\paragraph{3D Multimodal Large Language Models (MLLMs).}
Motivated by the success of 2D MLLMs~\cite{qwen2_5_vl, yang2025qwen3, gpt4o, llava_onevision, vila, llava, gemini_pro, cambirian1, qwen2_vl, yoon2025viral}, a growing body of work extends them to 3D understanding. Early efforts feed explicit 3D inputs to the LLM, either by lifting 2D features into 3D~\cite{llava3d, 3dllm} or by training dedicated point-cloud encoders as an additional modality~\cite{ll3da, wang2023chat, leo, pointllm}; the scarcity of paired 3D--language data, however, has limited their performance on language-heavy tasks. More recent approaches retain the multi-view 2D input format and inject 3D awareness through training objectives or auxiliary modules. One family supplies pixel-level 3D signals to encourage cross-view correspondence: Video-3D-LLM~\cite{video3dllm} and LLaVA-3D~\cite{llava3d} attach 3D positional embeddings derived from input point clouds, GPT4Scene~\cite{gpt4scene} renders bird's-eye-view markers, Ross3D~\cite{wang2025ross3d} adds a cross-view reconstruction pretext task, and 3DRS~\cite{huang20253drs} distills features with explicit correspondence supervision. A second family fuses MLLM image features with representations from 3D foundation models~\cite{wang2025continuous, wang2025vggt} to expose the LLM to dense, geometry-aware features, as in VLM3R~\cite{fan2025vlm}, $\text{G}^2$VLM~\cite{hu2025g2vlm}, and the concurrent 3DThinker~\cite{chen2025think}. Despite steady progress, a substantial gap to human-level performance on multi-view 3D reasoning benchmarks remains~\cite{vsibench}.\vspace{-5pt}

\paragraph{Compact Scene Representations.}
Cognitive studies suggest that humans do not maintain pixel-accurate reconstructions of every surface; rather, they rely on coarse abstractions that are nonetheless sufficient to support rich spatial reasoning~\cite{burgess2006spatial, shepard1971mental, lee2025perspective}. A line of work in 3D vision has correspondingly explored decomposing scenes into compact geometric primitives such as meshes, polygons, or superquadrics~\cite{tulsiani2017learning, paschalidou2019superquadrics, paschalidou2020learning, fedele2025superdec, monnier2023differentiable}, but these methods either struggle to scale beyond a handful of primitives~\cite{monnier2023differentiable} or rely on off-the-shelf 3D instance segmentation~\cite{fedele2025superdec}. More recently, C3G~\cite{an2025c3g} introduced a feed-forward pipeline that represents a scene compactly with 3D Gaussian Splatting~(3DGS) parameters, decoding 3D Gaussians from a small set of learnable query tokens supervised purely through photometric rendering. Inspired by this line of work, we bring token-level Gaussian decoding inside an MLLM, where a small set of \emph{Gaussian summary tokens} learns to assemble a compact 3D representation of the scene that the LLM conditions on when reasoning about it.\vspace{-5pt}
\section{Methodology}
\label{sec:method}
\subsection{Preliminaries: Multi-view MLLMs}
\label{subsec:preliminaries}
Following prior works~\cite{wang2025ross3d,video3dllm,llava3d}, we build our model on top of MLLMs capable of taking multiple images as input. Here, we briefly explain the input formulation of such MLLMs.

A multi-view MLLM is composed of a vision encoder $V_\psi(\cdot)$, a vision--language projector $P_\phi(\cdot)$, and a pre-trained large language model $LM_\theta(\cdot)$, where $\psi$, $\phi$, and $\theta$ denote the corresponding parameters. Given a set of $K$ input frames $\mathcal{I} = \{I_k\}_{k=1}^{K}$ with $I_k \in \mathbb{R}^{H \times W \times 3}$, the vision encoder independently extracts patch-level features from each frame:
\begin{equation}
    \mathbf{z}_k = V_\psi(I_k) \in \mathbb{R}^{N \times D_\mathbf{z}}, \quad k = 1, \dots, K,
\end{equation}
where $N$ is the number of patches per frame and $D_\mathbf{z}$ is the visual feature dimension. Each frame's features are then projected into the language model's embedding space of dimension $D$ via $P_\phi(\cdot)$, yielding per-frame visual tokens $\mathbf{e}^{\mathrm{img}}_k = P_\phi(\mathbf{z}_k) \in \mathbb{R}^{N' \times D}$, where $N'$ may differ from $N$ due to spatial token-reduction operations such as bilinear pooling and per-row newline insertion that preserve the 2D spatial layout~\cite{zhang2024llavavideo}. Concatenating all $K$ frames produces the full visual token sequence:
\begin{equation}
    \mathbf{e}^{\mathrm{img}} = [\mathbf{e}^{\mathrm{img}}_1; \mathbf{e}^{\mathrm{img}}_2; \dots; \mathbf{e}^{\mathrm{img}}_K] \in \mathbb{R}^{KN' \times D}.
\end{equation}
The input text (e.g.,~the user instruction) is tokenized and embedded into the same space through the language model's embedding layer, producing textual embeddings $\mathbf{e}^{\mathrm{text}} \in \mathbb{R}^{L \times D}$, where $L$ denotes the number of text tokens. The language model then takes the concatenated multimodal sequence $[\mathbf{e}^{\mathrm{img}}; \mathbf{e}^{\mathrm{text}}] \in \mathbb{R}^{(KN' + L) \times D}$ as input and models the causal distribution over the text tokens as:
\begin{equation}
    p_{\theta,\phi}(\mathbf{e}^{\mathrm{text}}_{1:L} \mid \mathbf{e}^{\mathrm{img}}) = \prod_{i=1}^{L} p_{\theta,\phi}\bigl(\mathbf{e}^{\mathrm{text}}_i \,\big|\, \mathbf{e}^{\mathrm{text}}_{<i},\, \mathbf{e}^{\mathrm{img}}\bigr).
\end{equation}
At inference, the language model autoregressively generates text tokens conditioned on the visual tokens and the previously generated text tokens. 

\begin{figure}[t]
    \centering
    \includegraphics[width=1\linewidth]{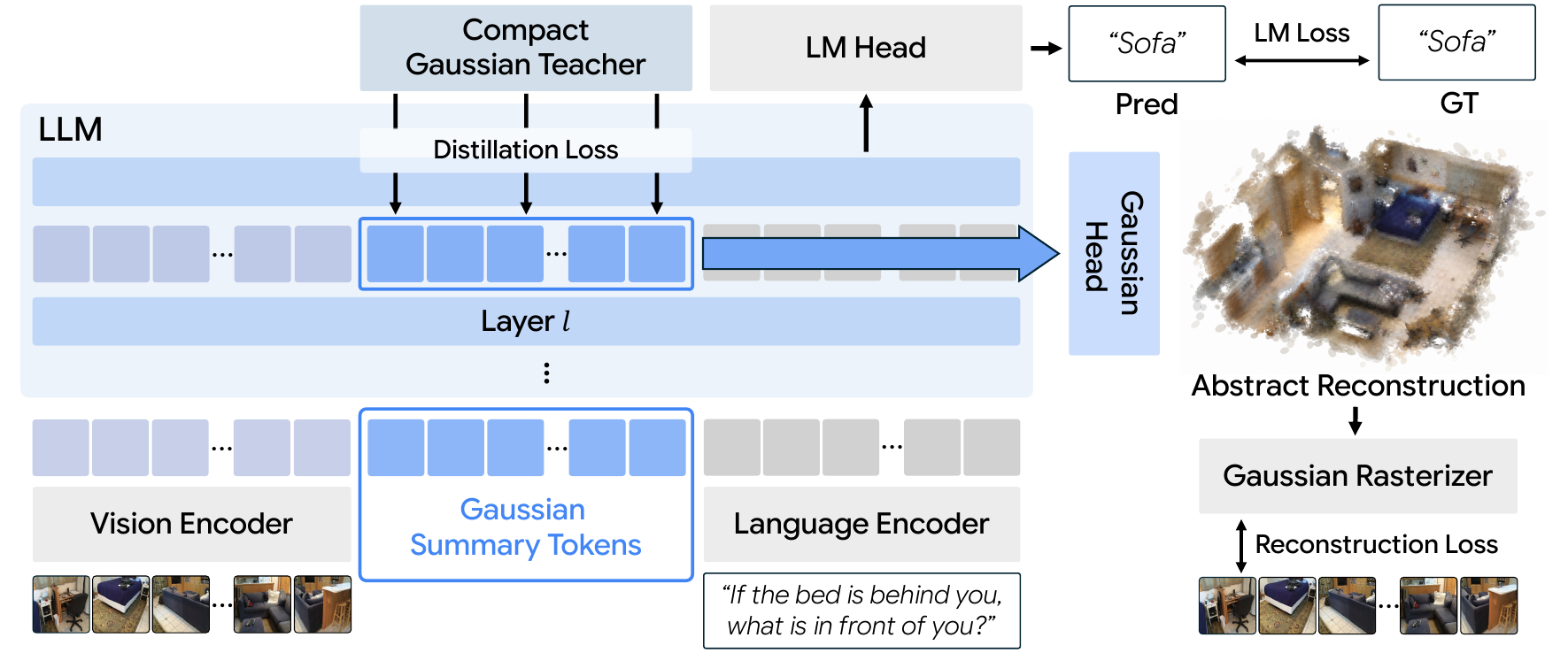}
    \vspace{-10 pt}
    \caption{\textbf{Overall architecture of \model.} Given multi-view images of a scene and a text instruction, our model appends a compact set of learnable \textcolor{g_sum}{\emph{Gaussian summary tokens}} between the image and text tokens. At an intermediate layer $\ell$, the hidden states at the Gaussian summary token positions are decoded into a compact 3D Gaussian Splatting representation that abstractly reconstructs the scene. The model is trained jointly with a standard language modeling loss on the text output, a photometric reconstruction loss obtained by rendering the predicted Gaussians at the input viewpoints, and a distillation loss from a pretrained compact Gaussian teacher~\cite{an2025c3g}.}
    \label{fig:architecture}
    \vspace{-10pt}
\end{figure}

\subsection{Imagine3D-LLM}
\label{subsec:method}
We now describe how we equip a multi-image MLLM with the ability to perform mental 3D reconstruction of the input scene for improved 3D reasoning. At a high level, we introduce a compact set of learnable \emph{Gaussian summary tokens} that are appended to the image tokens fed into the LLM (\S~~\ref{subsubsec:gaussian_tokens}). These tokens learn to summarize the multi-view observations into a compact set of 3D Gaussians, jointly trained with the reconstruction and standard language modeling objective. Since pure photometric supervision is known to require lengthy training schedules even for learning Gaussians alone~\cite{chen2024mvsplat, charatan2024pixelsplat,ye2024no,hong2024pf3plat,jiang2025anysplat,an2025c3g}, we further introduce a distillation objective from a pretrained Gaussian teacher model~\cite{veicht2026zipsplat}, to substantially accelerate convergence (\S~\ref{subsubsec:distillation}). Finally, we analyze how this joint training improves the MLLM's internal representations~(\S~\ref{subsec:analysis}). An overview of our method is shown in Fig.~\ref{fig:architecture}.

\subsubsection{Mental Reconstruction via Gaussian Summary Tokens}
\label{subsubsec:gaussian_tokens}

\paragraph{Gaussian summary tokens.}
Given the visual token sequence $\mathbf{e}^{\mathrm{img}} \in \mathbb{R}^{KN' \times D}$ produced by the vision encoder and projector, we introduce a compact set of $M$ learnable Gaussian summary tokens $\mathbf{e}^{\mathrm{gs}} \in \mathbb{R}^{M \times D}$, where $M \ll KN'$. These tokens are inserted between the visual tokens and the textual tokens, yielding the augmented multimodal sequence
\begin{equation}
    [\mathbf{e}^{\mathrm{img}};\, \mathbf{e}^{\mathrm{gs}};\, \mathbf{e}^{\mathrm{text}}] \in \mathbb{R}^{(KN' + M + L) \times D},
\end{equation}
which is processed by the LLM via standard causal self-attention. By placing the Gaussian summary tokens \emph{after} the image tokens, each token can attend to all visual evidence and selectively retrieve the information needed to faithfully describe a portion of the 3D scene, rather than being passively bound to local image content.\vspace{-5pt}

\paragraph{Bottleneck-induced object grouping.}
A central design choice is that the number of Gaussian summary tokens is substantially smaller than the number of image tokens, $M \ll KN'$. This bottleneck prevents the model from naively allocating one token per pixel or per patch, and instead forces overlapping content observed across multiple views to be merged into a shared set of tokens. Under this constraint, the most efficient way for the model to faithfully reconstruct the scene with limited Gaussians is to assign each token to a coherent 3D region or object that recurs across views. As we show in \S~\ref{subsec:analysis}, this bottleneck gives rise to an emergent object-centric grouping, mirroring the way humans build object-level mental abstractions of a scene before reasoning about it. Since our main objective is to improve 3D reasoning rather than novel view synthesis, this design also offers a favorable trade-off between reconstruction quality and computational cost, preventing the input sequence length to increase substantially.\vspace{-5pt}

\paragraph{Decoding 3D Gaussians.}
Given the LLM forward pass over the augmented sequence $[\mathbf{e}^{\mathrm{img}};\, \mathbf{e}^{\mathrm{gs}};\, \mathbf{e}^{\mathrm{text}}]$, a natural question is from which layer the Gaussian summary tokens should be decoded. Recent analyses~\cite{kaduri2025whats, zhang2025cross, jiang2025devils, kang2025your, yoon2025viral} consistently identify that the \emph{middle layers} of the LLM most actively show information flow between visual and text modalities. Since our Gaussian summary tokens also require effectively absorbing visual information from the image tokens, we extract the hidden states at the Gaussian summary token positions from the middle layer $\ell$, denoted $\mathbf{h}^{\mathrm{gs}} \in \mathbb{R}^{M \times D}$. Decoding at a middle layer also leaves enough subsequent layers for the text tokens to attend to the Gaussian summary tokens, enabling the text tokens to benefit from the scene-summarizing information encoded by the Gaussian tokens. The effectiveness of choosing $\ell$ as a middle layer is further validated in Tab.~\ref{tab:abl_layer}.

The extracted hidden states $\mathbf{h}^{\mathrm{gs}}$ are further decoded into 3D Gaussian primitives through a lightweight MLP-based Gaussian head $\mathcal{H}(\cdot)$. Rather than mapping each token to a single Gaussian, we let each token decode $G$ Gaussians simultaneously, providing additional representational capacity per token without expanding the LLM's sequence length:
\begin{equation}
    \{\mathbf{G}_{i,j}\}_{j=1}^{G} = \mathcal{H}(\mathbf{h}^{\mathrm{gs}}_i), \quad i = 1, \dots, M,
\end{equation}
yielding a total of $M \times G$ Gaussians per scene. Each Gaussian $\mathbf{G}_{i,j} = \{\mu_{i,j}, \sigma_{i,j}, \Sigma_{i,j}, c_{i,j}\}$ is parameterized by its 3D center $\mu \in \mathbb{R}^3$, opacity $\sigma \in [0, 1)$, covariance matrix $\Sigma \in \mathbb{R}^{3 \times 3}$, and spherical harmonics coefficients $c \in \mathbb{R}^{3(L_{\mathrm{SH}}+1)^2}$ that encode view-dependent color with $L_{\mathrm{SH}}$ degrees. Together, the resulting set $\{\mathbf{G}_{i,j}\}$ forms a compact 3D representation of the scene that can be rendered into novel views via differentiable rasterization~\cite{kerbl20233d}.

\paragraph{Training objective.}
We jointly train the model with two losses. The standard \emph{language modeling loss} $\mathcal{L}_{\mathrm{LM}}$ supervises the autoregressive prediction of text tokens conditioned on the visual and Gaussian summary tokens, while the \emph{reconstruction loss} $\mathcal{L}_{\mathrm{recon}}$ supervises the decoded Gaussians by rendering them at the input viewpoints $\{\pi_k\}_{k=1}^{K}$ via differentiable rasterization and comparing the resulting images to the ground-truth frames $I_k$:
\begin{align}
\mathcal{L}_{\mathrm{LM}} &= -\sum_{i=1}^{L} \log p_{\theta,\phi}\bigl(\mathbf{e}^{\mathrm{text}}_i \,\big|\, \mathbf{e}^{\mathrm{text}}_{<i},\, \mathbf{e}^{\mathrm{img}},\, \mathbf{e}^{\mathrm{gs}}\bigr), \\
\mathcal{L}_{\mathrm{recon}} &= \sum_{k=1}^{K} \Bigl[ \lambda_{\mathrm{MSE}} \mathcal{L}_{\mathrm{MSE}}(\hat{I}_k, I_k) + \lambda_{\mathrm{LPIPS}} \mathcal{L}_{\mathrm{LPIPS}}(\hat{I}_k, I_k) \Bigr],
\end{align}
where $\hat{I}_k$ is the image rendered from $\{\mathbf{G}_i\}_{i=1}^{M}$ at viewpoint $\pi_k$. The combined objective is $\mathcal{L} = \mathcal{L}_{\mathrm{LM}} + \lambda_{\mathrm{recon}} \mathcal{L}_{\mathrm{recon}}$.

\subsubsection{Accelerating Convergence via Distillation from a Compact Gaussian Teacher}
\label{subsubsec:distillation}
Although the model can in principle be trained end-to-end with the objective above, we find that pure photometric supervision through the LLM converges slowly, requiring substantially more iterations than is practical given the cost of MLLM finetuning. This is consistent with prior observations in the feed-forward 3D Gaussian Splatting literature, where dedicated estimators are typically trained for hundreds of thousands of iterations with large batch sizes to obtain reliable Gaussian predictions~\cite{chen2024mvsplat, charatan2024pixelsplat,ye2024no,hong2024pf3plat,jiang2025anysplat,an2025c3g}. To circumvent this bottleneck, we leverage a pretrained compact Gaussian estimator as a teacher model and distill its scene representation into the LLM, providing a strong inductive signal for what the Gaussian summary tokens should encode.

\paragraph{Compact Gaussian teacher.}
We adopt ZipSplat~\cite{veicht2026zipsplat}, a recent feed-forward 3DGS framework that estimates a compact set of 3D Gaussians from multi-view images, as our teacher. Given the same multi-view inputs $\mathcal{I}$, ZipSplat first encodes them with a geometry-grounded visual backbone~\cite{wang2025vggt}, and then refines a small set of $M$ learnable query tokens $\mathbf{Q}^{\mathrm{T}} \in \mathbb{R}^{M \times d_{\mathrm{T}}}$ through a transformer that jointly attends over the queries and the multi-view features, where $d_{\mathrm{T}}$ denotes the teacher model's hidden dimension. The refined tokens $\bar{\mathbf{Q}}^{\mathrm{T}} \in \mathbb{R}^{M \times d_{\mathrm{T}}}$ are then decoded into 3D Gaussians $\{\mathbf{G}^{\mathrm{T}}_{i,j}\}_{j=1}^{G}$ for $i = 1, \dots, M$ through a lightweight Gaussian head. Crucially, we match the number of teacher tokens $M$ in ZipSplat's inference time with the number of Gaussian summary tokens in our LLM, allowing direct token-level alignment between teacher and student.\vspace{-5pt}

\paragraph{Distillation loss.}
We distill from the teacher at two complementary levels. First, we align the LLM's hidden states at the Gaussian summary token positions, $\mathbf{h}^{\mathrm{gs}} \in \mathbb{R}^{M \times D}$, with the teacher's refined query tokens $\bar{\mathbf{Q}}^{\mathrm{T}}$ via a normalized cosine similarity loss with a lightweight projection layer $g(\cdot)$ that maps to the teacher's dimension $d_{\mathrm{T}}$, $\mathcal{L}_{\mathrm{token}} = \frac{1}{M} \sum_{i=1}^{M} \left[ 1 - \cos\!\left( \frac{g(\mathbf{h}^{\mathrm{gs}}_i)}{\|g(\mathbf{h}^{\mathrm{gs}}_i)\|},\; \frac{\bar{\mathbf{Q}}^{\mathrm{T}}_i}{\|\bar{\mathbf{Q}}^{\mathrm{T}}_i\|} \right) \right]$,
where $\cos(\cdot, \cdot)$ denotes the cosine similarity operation and $\|\cdot\|$ is the L2-norm. This token-level supervision provides a dense, per-token target that guides the LLM toward representations the teacher's Gaussian head can already decode, bypassing the slow convergence of learning Gaussian-decodable features from photometric loss alone. Second, we additionally supervise the student's decoded Gaussian parameters against the teacher's predictions with an L2 loss over all Gaussian attributes $\mathcal{L}_{\mathrm{param}} = \frac{1}{M} \sum_{i=1}^{M} \bigl\| \mathbf{G}_i - \mathbf{G}^{\mathrm{T}}_i \bigr\|_2^2$,
where $\mathbf{G}_i$ and $\mathbf{G}^{\mathrm{T}}_i$ are taken as the concatenation of all predicted Gaussian attributes from the student and teacher, respectively. The total distillation loss is $\mathcal{L}_{\mathrm{distill}} = \lambda_{\mathrm{token}} \mathcal{L}_{\mathrm{token}} + \lambda_{\mathrm{param}} \mathcal{L}_{\mathrm{param}}$.\vspace{-5pt}

\paragraph{Final objective.}
Combining the language modeling, reconstruction, and distillation losses, the full training objective of \model\ is:
\begin{equation}
    \mathcal{L}_{\text{full}} = \mathcal{L}_{\mathrm{LM}} + \lambda_{\mathrm{recon}} \mathcal{L}_{\mathrm{recon}} + \lambda_{\mathrm{distill}} \mathcal{L}_{\mathrm{distill}}.
\end{equation}
The teacher is kept frozen throughout training and is only used at training time.

\subsection{Analysis}
\label{subsec:analysis}
In this section, we further analyze the characteristics of the trained model to understand how jointly training the model to estimate coarse reconstructions of the scene leads to improved 3D reasoning.

\begin{figure}[t]
    \centering
    \includegraphics[width=1\linewidth]{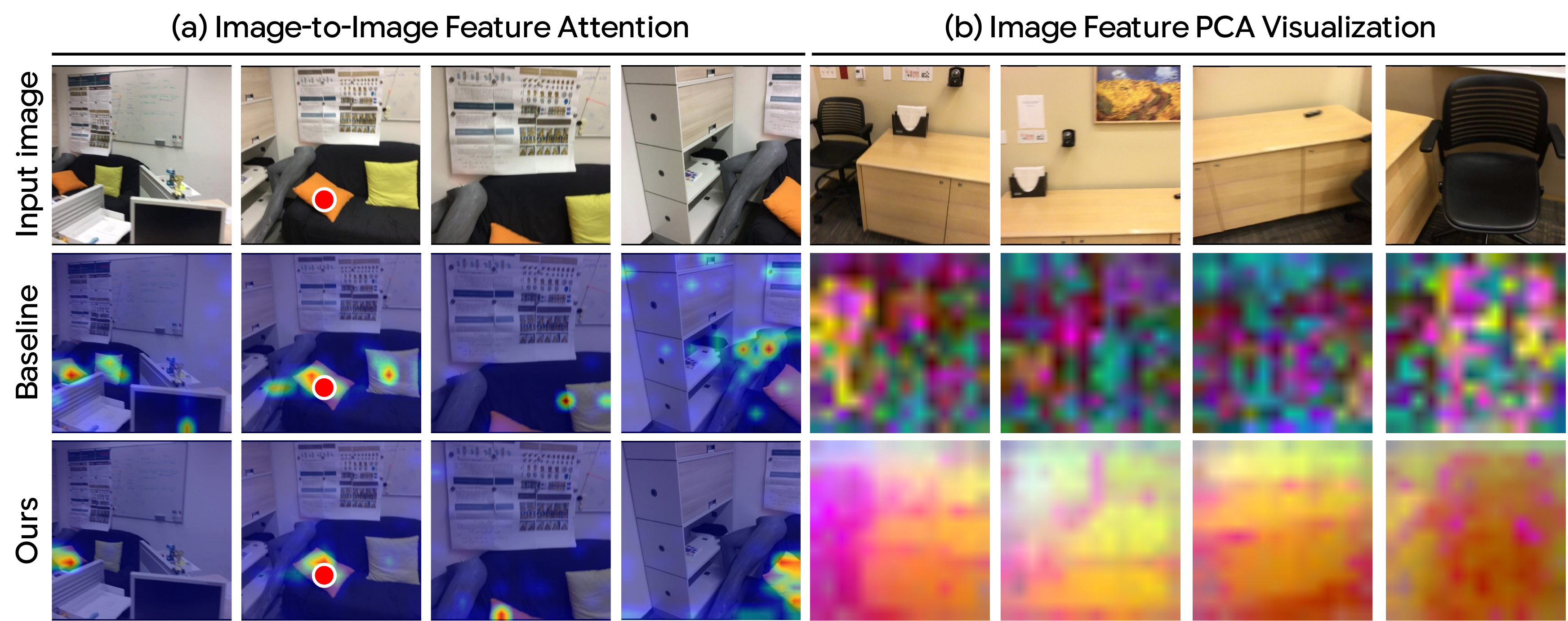}
    \vspace{-10pt}
    \caption{\textbf{Effects of joint reconstruction training on the LLM's image features.} \textbf{(a)} Image-to-image attention maps between a query point (\textcolor{red}{red} dot) and other input views. Our model exhibits substantially stronger cross-view correspondences than the baseline, indicating that the Gaussian summary tokens implicitly drive the image features to align across views. \textbf{(b)} PCA visualization of the LLM's image features (first three components shown as RGB). Our features are noticeably cleaner and more semantically structured, with corresponding objects (e.g., chairs, desks) encoded with consistent colors across views, reflecting the emergence of object-level abstractions.}
    \label{fig:corr_pca}
    \vspace{-10pt}
\end{figure}

\paragraph{Joint training improves image feature representations.}
\label{subsubsec:feature_quality}
While our reconstruction loss is applied only at the Gaussian summary tokens and does not directly supervise the image features, we observe that the LLM's image features themselves become substantially more 3D-aware after joint training. Fig.~\ref{fig:corr_pca}-(a) visualizes the attention maps between a query point in one view~(\textcolor{red}{red} dot) and the image features of the other input views. Compared to the baseline, our model produces noticeably sharper and more spatially focused attention on the corresponding region across views, indicating that the same object is now encoded with consistent representations regardless of viewpoint. Beyond this qualitative observation, we additionally evaluate cross-view correspondence accuracy of our model's image features throughout training, and observe a steady improvement as training progresses~(Appendix~\ref{appendix:correspondence}). Fig.~\ref{fig:corr_pca}-(b) makes the underlying structure more explicit by visualizing the first three principal components of the image-token hidden states at the $\ell$-th LLM layer in RGB. While the baseline features appear noisy and largely unstructured, our features are markedly cleaner and more semantically organized, with corresponding regions encoded with consistent colors across viewpoints. Together, these two effects indicate that the reconstruction objective propagates 3D-aware structure beyond the Gaussian summary tokens into the LLM's image features themselves, encouraging the same kind of compact, view-consistent representation that the summary tokens are trained to produce.\vspace{-5pt}

\begin{figure}[t]
    \centering
    \includegraphics[width=1\linewidth]{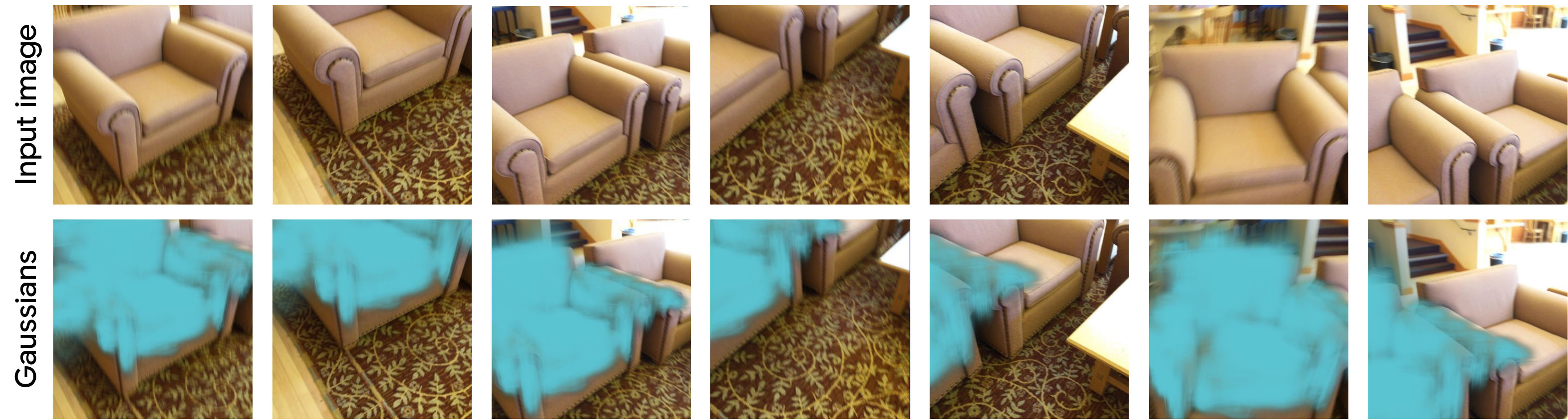}
    \vspace{-10 pt}
    \caption{\textbf{Visualization of clustered Gaussian from summary tokens.} We visualize the Gaussians clustered from a set of Gaussian summary tokens via K-means over the summary tokens (\textcolor{cyan}{cyan} indicates the Gaussians belonging to one cluster). Similar summary tokens are mapped to the same object without any explicit clustering supervision.}
    \label{fig:g2i}
    \vspace{-15pt}
\end{figure}
\paragraph{Summary tokens emergently group by object.}
\label{subsubsec:emergent_grouping}
A central design choice of \model\ is the bottleneck $M \ll KN'$, which compels overlapping content across views to be merged into a shared set of summary tokens~(\S\ref{subsubsec:gaussian_tokens}). To examine whether this bottleneck actually induces object-level grouping, we cluster the trained summary tokens via K-means and visualize, for each cluster, the 3D Gaussians decoded from its member tokens. As shown in Fig.~\ref{fig:g2i}, a single cluster's Gaussians reliably localize one object (e.g.,~the chair) across all views, despite the model receiving no clustering supervision. This emergent grouping effectively validates that learning to reconstruct the scene through a small set of summary tokens naturally guides the model toward the abstract, object-level understanding our design aims to achieve, without requiring any explicit object-level supervision.

\section{Experiments}\label{sec:exp}

\paragraph{\textbf{Implementation Details.}}
\label{subsec:imde}
Following prior work~\cite{video3dllm,wang2025ross3d}, we build our \model\ based on LLaVA-Video-7B~\cite{zhang2024llavavideo}. LLaVA-Video encodes each image as 210 tokens, resulting in $N'=210$ and $KN'=6720$ tokens for most settings where we use 32 images as input. For 3DGS estimation, we choose $\ell=14$, as it is the middle layer of the base LLM. For the number of Gaussian summary tokens, we set $M=2592$, which corresponds to 81 tokens per input image and is much fewer than the total number of image tokens $KN'=6720$. The number of Gaussians decoded per token is set as $G=32$. The Gaussian decoder head $\mathcal{H}(\cdot)$ is a three-layer MLP. We fine-tune the entire model on the combination of ScanNet-based datasets~(SQA3D~\cite{ma2022sqa3d}, ScanQA~\cite{scanqa}, Scan2Cap~\cite{scan2cap}, ScanRefer~\cite{scanrefer}, and Multi3DRefer~\cite{zhang2023multi3drefer}). Motivated by recent analysis~\cite{zhang2025scaling,zhou2025vlm4d, zheng2025learning} where training only with ScanNet-based datasets show largely overfitted results to ScanNet rather than unlocking 3D reasoning, we also sample and add 100K subset from SPAR-7M~\cite{spar}. The training is conducted with 16 GH200 GPUs, with an effective batch size of 64 for 1 epoch. More detailed implementation details can be found in Appendix~\ref{appendix:implementation_detail}.\vspace{-5pt}

\paragraph{\textbf{Datasets and metrics.}} Following prior work~\cite{video3dllm,wang2025ross3d, zheng2025learning, hu2025g2vlm}, we evaluate on seven benchmarks. Six are from ScanNet~\cite{dai2017scannet}: SQA3D~\cite{ma2022sqa3d}, Real-3DQA~\cite{ma2026real3dqa}, and ScanQA~\cite{scanqa} for situated and spatial reasoning, Scan2Cap~\cite{scan2cap} for 3D grounded captioning, and ScanRefer~\cite{scanrefer} and Multi3DRefer~\cite{zhang2023multi3drefer} for 3D object detection. We additionally evaluate on SPAR-Bench~\cite{spar}, which assesses genuine 3D perception and reasoning across three cognitive levels: low-level (e.g., depth estimation), medium-level (e.g., view-change inference), and high-level (e.g., spatial imagination). Following prior work~\cite{llava3d,video3dllm,wang2025ross3d,zheng2025learning,fan2025vlm,hu2025g2vlm}, we use exact / refined-exact match (EM / EM-R)~\cite{leo,llava3d,ma2026real3dqa} on SQA3D and Real-3DQA, EM together with BLEU-4~\cite{papineni2002bleu}, METEOR~\cite{banerjee2005meteor}, ROUGE~\cite{lin2004rouge}, and CIDEr~\cite{vedantam2015cider} on ScanQA, the same captioning metrics under IoU@0.5 on Scan2Cap, Acc@0.25/0.5 on ScanRefer and the corresponding F1 on Multi3DRefer, and accuracy per cognitive level with the overall mean on SPAR-Bench.

\begin{table*}[t]
    \centering\small
    \caption{
\textbf{Quantitative results on 3D question-answering} over SQA3D~\cite{ma2022sqa3d}, Real-3DQA~\cite{ma2026real3dqa}, and ScanQA~\cite{scanqa}.
    ``Specialists'' refer to models tailored to individual tasks via task-specific decoders.
    General-purpose 2D MLLMs~\cite{internvl2, qwen2_vl, zhang2024llavavideo} are reported under a zero-shot protocol. We report values available from prior works; ``--'' marks entries unavailable to us.}
    \label{tab:3dqa}
    \resizebox{\linewidth}{!}{
    \begin{tabular}{l cc cc ccccc}
    \toprule
    \multirow{2}{*}{Method} & \multicolumn{2}{c}{SQA3D$_{\text{test}}$} & \multicolumn{2}{c}{Real-3DQA$_{\text{test}}$} & \multicolumn{5}{c}{ScanQA$_{\text{val}}$} \\
    \cmidrule(lr){2-3} \cmidrule(lr){4-5} \cmidrule(lr){6-10}
    & EM & EM-R & EM & EM-R & CIDEr & BLEU-4 & METEOR & ROUGE & EM \\
    \midrule
    \multicolumn{10}{l}{\textit{Specialists}} \\
    SQA3D~\cite{ma2022sqa3d}                  & 46.6 & --   & --   & --   & --    & --   & --   & --   & --   \\
    ScanQA~\cite{scanqa}                      & --   & --   & --   & --   & 64.9  & 10.1 & 13.1 & 33.3 & 21.1 \\
    \midrule
    \multicolumn{10}{l}{\textit{2D MLLMs}} \\
    InternVL2-8B~\cite{internvl2}             & 33.0 & 45.3 & --   & --   & 62.5  & 3.3  & 14.5 & 34.3 & --   \\
    Qwen2-VL-7B~\cite{qwen2_vl}               & 40.7 & 46.7 & --   & --   & 53.9  & 3.0  & 11.4 & 29.3 & --   \\
    LLaVA-Video-7B~\cite{zhang2024llavavideo} & 48.5 & --   & --   & --   & 88.7  & 3.1  & 17.7 & 44.6 & --   \\
    \midrule
    \multicolumn{10}{l}{\textit{3D MLLMs}} \\
    Scene-LLM~\cite{scenellm}                 & 53.6 & --   & --   & --   & 80.0  & 11.7 & 15.8 & 35.9 & 27.2 \\
    LL3DA~\cite{ll3da}                        & --   & --   & --   & --   & 76.8  & --   & 15.9 & 37.3 & --   \\
    LEO~\cite{leo}                            & 50.0 & 52.4 & --   & --   & 80.0  & 11.5 & 16.2 & 39.3 & 21.5 \\
    ChatScene~\cite{huang2024chatscene}       & 54.6 & 57.5 & --   & --   & 87.7  & 14.3 & 18.0 & 41.6 & 21.6 \\
    Grounded 3D-LLM~\cite{grounded-3dllm}     & --   & --   & --   & --   & 72.7  & 13.4 & --   & --   & --   \\
    LLaVA-3D~\cite{llava3d}                   & 55.6 & 57.6 & --   & --   & 91.7  & 14.5 & 20.7 & 50.1 & 27.0 \\
    Video-3D-LLM~\cite{video3dllm}            & 58.6 & --   & 31.2   & 36.4   & 102.1 & 16.4 & 20.0 & 49.3 & 30.1 \\
    3DRS~\cite{huang20253drs}            & 60.6 & --   & -   & -   & 104.8 & - & - & - & 30.3 \\
    Ross3D~\cite{wang2025ross3d}              & 63.0 & 65.7 & 36.6   & 41.5   & 107.0 & 17.9 & 20.9 & \textbf{50.7} & \textbf{30.8} \\
    \rowcolor{Light!15}
    \textbf{\model~(Ours)}                    & \textbf{63.8}   & \textbf{66.4}   & \textbf{39.2} & \textbf{44.3}   & \textbf{109.3}    & \textbf{20.5}   & \textbf{21.3}   & 49.7   & 29.9   \\
    \bottomrule
    \end{tabular}
    }
\end{table*}

\begin{table*}[t]
    \centering\small
    \caption{
    \textbf{Evaluation of 3D dense captioning and visual grounding} on Scan2Cap~\cite{scan2cap}, ScanRefer~\cite{scanrefer}, and Multi3DRefer~\cite{zhang2023multi3drefer}.
    }
    \label{tab:caption_grounding}
    \resizebox{\textwidth}{!}{%
    \begin{tabular}{l cccc cc cc}
    \toprule
    \multirow{2}{*}{Method} & \multicolumn{4}{c}{Scan2Cap$_{\text{val}}$ (IoU@0.5)} & \multicolumn{2}{c}{ScanRefer$_{\text{val}}$} & \multicolumn{2}{c}{Multi3DRefer$_{\text{val}}$} \\
    \cmidrule(lr){2-5} \cmidrule(lr){6-7} \cmidrule(lr){8-9}
    & ROUGE & BLEU-4 & METEOR & CIDEr & Acc@0.25 & Acc@0.5 & F1@0.25 & F1@0.5 \\
    \midrule
    \multicolumn{9}{l}{\textit{Specialists}} \\
    Scan2Cap~\cite{scan2cap}          & 44.5 & 23.3 & 22.0 & 35.2          & --   & --   & --   & --   \\
    3DJCG~\cite{3djcg}                 & 50.8 & 31.0 & 24.2 & 49.5          & 49.6 & 37.3 & --   & 26.6 \\
    ScanRefer~\cite{scanrefer}        & --   & --   & --   & --            & 37.3 & 24.3 & --   & --   \\
    M3DRef-CLIP~\cite{zhang2023multi3drefer}  & --   & --   & --   & --            & 51.9 & 44.7 & 42.8 & 38.4 \\
    \midrule
    \multicolumn{9}{l}{\textit{3D MLLMs}} \\
    LL3DA~\cite{ll3da}                & 55.1 & 36.8 & 26.0 & 65.2          & --   & --   & --   & --   \\
    Ground 3D-LLM~\cite{grounded-3dllm}     & --   & --   & --   & --            & 47.9 & 44.1 & 45.2 & 40.6 \\
    LEO~\cite{leo}                   & 58.1 & 38.2 & 27.9 & 72.4          & --   & --   & --   & --   \\
    ChatScene~\cite{huang2024chatscene}       & 58.1 & 36.3 & --   & 77.1          & 55.5 & 50.2 & 57.1 & 52.4 \\
    LLaVA-3D~\cite{llava3d}            & 63.4 & 41.1 & 30.2 & 79.2          & 54.1 & 42.4 & --   & --   \\
    Video-3D-LLM~\cite{video3dllm}   & 62.3 & 42.4 & 28.9 & 83.8 & 58.1 & 51.7 & 58.0 & 52.7 \\
    VG-LLM~\cite{zheng2025learning}   & 62.6 & 41.5 & 28.9 & 80.0 & - & - & - & - \\
    3DRS~\cite{huang20253drs}   & - & 41.6 & - & 86.1 & - & - & - & - \\
    Ross3D~\cite{wang2025ross3d} & 66.9 & 43.4 & 30.3 & 81.3 & 61.1 & 54.4 & 59.6 & 54.3 \\
    \rowcolor{Light!15}
    \textbf{\model~(Ours)} & \textbf{67.6} & \textbf{48.3} & \textbf{32.6} & \textbf{99.2} & \textbf{62.8} & \textbf{56.3} & \textbf{60.2} & \textbf{55.0} \\
    \bottomrule
    \end{tabular}%
    }
\end{table*}

\subsection{Quantitative Comparison}
\label{sec:comparison}

\paragraph{\textbf{ScanNet-based datasets.}}
On the ScanNet-based benchmarks (SQA3D~\cite{ma2022sqa3d}, Real-3DQA~\cite{ma2026real3dqa}, ScanQA~\cite{scanqa}, Scan2Cap~\cite{scan2cap}, ScanRefer~\cite{scanrefer}, and Multi3DRefer~\cite{zhang2023multi3drefer}), we compare \model\ against task-specific specialists, general-purpose 2D LMMs, and recent 3D LMMs that differ in how they encode the 3D scene. As shown in Tab.~\ref{tab:3dqa}, and Tab.~\ref{tab:caption_grounding}, \model\ significantly improves performance across benchmarks, outperforming the strongest prior 3D LMM (Ross3D) on almost all benchmarks. Especially, state-of-the-art performance in Real-3DQA is notable, which is a benchmark explicitly designed to validate 3D reasoning capabilities and penalize language-shortcut overfitting~\cite{ma2026real3dqa}, indicating that our improvements stem from genuine 3D reasoning.\vspace{-5pt}

\paragraph{SPAR-Bench.}
On SPAR-Bench~\cite{spar}, we compare \model\ against three categories of models: the \textit{proprietary} group~\cite{gpt4o, anthropic2025claude37sonnet}, the \textit{general-purpose 2D LMMs} group~\cite{zhang2024llavavideo,qwen2_vl,internvl2,qwen2_5_vl}, and the \textit{spatially-aware} group of recent models specifically designed for spatial reasoning~\cite{fan2025vlm,hu2025g2vlm,chen2025think}. 
\begin{wraptable}[19]{r}{0.5\linewidth}
    \centering\footnotesize
    % \vspace{-30pt}
    \setlength{\tabcolsep}{3pt}
    \caption{\textbf{Evaluation on SPAR-Bench}~\cite{spar}.}
    \label{tab:spar}
    \resizebox{\linewidth}{!}{%
    \begin{tabular}{l cccc}
    \toprule
    \multirow{2}{*}{Model} & \multicolumn{4}{c}{SPAR-Bench} \\
    \cmidrule(lr){2-5}
    & Avg. & Low & Med. & High \\
    \midrule
    \multicolumn{5}{l}{\textit{Proprietary}} \\
    GPT-4o~\cite{gpt4o} & 36.4 & 29.3 & 24.9 & 45.1 \\
    Claude-3.7-Sonnet~\cite{anthropic2025claude37sonnet} & 21.8 & 25.4 & 7.3 & 23.3 \\
    \midrule
    \multicolumn{5}{l}{\textit{General-purpose 2D MLLMs}} \\
    LLaVA-Video-7B~\cite{zhang2024llavavideo} & 32.3 & 23.6 & 24.8 & 42.6 \\
    Qwen2-VL-7B~\cite{qwen2_vl} & 30.7 & 27.5 & 20.4 & 37.0 \\
    InternVL2.5-8B~\cite{internvl2.5} & 36.3 & 29.5 & 31.9 & 43.8 \\
    Qwen2.5-VL-72B~\cite{qwen2_5_vl} & 39.4 & 35.4 & 23.1 & 48.4 \\
    \midrule
    \multicolumn{5}{l}{\textit{Spatially-aware}} \\
    VLM3R-7B~\cite{fan2025vlm} & 43.2 & 39.8 & 28.4 & 51.2 \\
    $\text{G}^2\text{VLM}$-SR-2B~\cite{hu2025g2vlm} & 54.9 & 60.0 & 36.3 & 56.5 \\
    3DThinker-7B~\cite{chen2025think} & 63.3 & - & - & - \\
    \midrule
    \rowcolor{Light!15}
    \textbf{\model-7B~(Ours)} & \textbf{68.5} & \textbf{60.5} & \textbf{67.0} & \textbf{76.0} \\
    \bottomrule
    \end{tabular}}
\end{wraptable}
As shown in Tab.~\ref{tab:spar}, \model\ achieves state-of-the-art performance on SPAR-Bench, outperforming all competing methods across every cognitive level by substantial margins. Compared to general-purpose 2D LMMs, \model\ surpasses even the largest 72B-scale model~\cite{qwen2_5_vl} by over 29 points despite using a 7B backbone. More notably, \model\ outperforms the strongest spatially-aware baseline, 3DThinker-7B~\cite{chen2025think}, by 5.2 points overall, with even larger margins over $\text{G}^2\text{VLM}$-SR-2B~\cite{hu2025g2vlm} and VLM3R-7B~\cite{fan2025vlm}. The improvement is especially pronounced on the medium- and high-level splits, which most directly require the object-level cross-view reasoning our reconstruction objective targets. These results validate our central claim: equipping MLLMs to assemble a compact 3D representation of the scene before answering substantially improves 3D reasoning.

\subsection{Ablation Studies}\label{sec:ablation}
\paragraph{\textbf{Controlled comparison with baseline.}} 
\begin{wraptable}[13]{r}{0.5\linewidth}
    \centering\small
    \vspace{-16pt}
    \caption{\textbf{Controlled comparison against the base MLLM.} Both models share the identical backbone, training data, and schedule; the baseline is trained with standard visual instruction tuning, without the Gaussian summary tokens or the reconstruction and distillation objectives.}
    \label{tab:baseline_controlled}
    \setlength{\tabcolsep}{4pt}
    \resizebox{\linewidth}{!}{%
    \begin{tabular}{l ccc}
    \toprule
    Method & SQA3D & ScanQA & Scan2Cap \\
    \midrule
    Baseline & 56.5 & 26.2 & 63.1 \\
    \rowcolor{Light!15}
    \textbf{\model~(Ours)} & \textbf{63.8} & \textbf{29.9} & \textbf{67.6} \\
    \midrule\midrule
    Method & ScanRefer & Multi3DRefer & SPAR \\
    \midrule
    Baseline & 58.3 & 57.4 & 60.9 \\
    \rowcolor{Light!15}
    \textbf{\model~(Ours)} & \textbf{62.8} & \textbf{60.2} & \textbf{68.5} \\
    \bottomrule
    \end{tabular}}
\end{wraptable}
To isolate the contribution of our mental-reconstruction objective from the effect of the training data and recipe, we compare \model\ against a baseline that shares the identical backbone, training data, and schedule but omits the Gaussian summary tokens together with the reconstruction and distillation objectives, i.e.,~standard visual instruction tuning. Here, we report one representative metric per benchmark EM score for SQA3D / ScanQA, ROUGE for Scan2Cap, Acc@25 for ScanRefer, F1@0.25 for Multi3DRefer, and average performance SPAR. As shown in Tab.~\ref{tab:baseline_controlled}, \model\ improves over this controlled baseline on every benchmark, confirming that the gains arise from our objective rather than from the additional training data alone. The improvement is especially pronounced on the medium-level cross-view split of SPAR-Bench~(+15.5, from 51.5 to 67.0; see Tab.~\ref{tab:spar}), which most directly exercises the object-level cross-view reasoning our reconstruction objective targets.

\paragraph{\textbf{Layer selection for Gaussian decoding.}}
\begin{wraptable}[8]{r}{0.4\linewidth}
\vspace{-12pt}
    \centering\footnotesize
    \caption{
    \textbf{Effect of the layer $\ell$ from which Gaussian tokens are decoded.}
    }
    \label{tab:abl_layer}
    % \vspace{-5pt}
    \begin{tabular}{l ll}
    \toprule
    Layer $\ell$ & $\text{SQA3D}_{\text{test}}$ & SPAR-Bench \\
    \midrule
    7 (early) & 60.7 & 60.2 \\
    \rowcolor{Light!15}
    14 (middle, Ours) & \textbf{63.8} & \textbf{68.5} \\
    21 (late) & 61.7 & 64.0 \\
    \bottomrule
    \end{tabular}
\end{wraptable}
We further ablate the choice of the LLM layer $\ell$ from which the Gaussian summary tokens are extracted and decoded into 3D Gaussians. 
As discussed in \S~\ref{subsubsec:gaussian_tokens}, we hypothesize that the middle layer of the LLM offers the best balance between two opposing requirements: the Gaussian tokens must accumulate enough visual information to faithfully describe the scene, yet the remaining layers must be sufficient to propagate this 3D-aware information into the text tokens for reasoning. We empirically verify this by extracting from layer $\ell \in \{7, 14, 21\}$ of the 28-layer LLM. As shown in Tab.~\ref{tab:abl_layer}, the middle layer ($\ell=14$) consistently performs best, validating our design choice and aligning with prior findings on information flow in MLLMs~\cite{kaduri2025whats, jiang2025devils, yoon2025viral}.

\clearpage

\paragraph{\textbf{Effectiveness of distillation.}}

\begin{wraptable}[11]{r}{0.5\linewidth}
    \centering\footnotesize
    % \vspace{-33pt}
    \caption{\textbf{Effect of distillation from the compact Gaussian teacher.}}
    \label{tab:abl_distill}
    \vspace{-3pt}
    \resizebox{\linewidth}{!}{%
    \begin{tabular}{l cc}
    \toprule
    Method & $\text{SQA3D}_{\text{test}}$ & SPAR-Bench \\
    \midrule
    Baseline (1 ep.) & 56.5 & 60.9 \\
    Baseline (2 ep.) & 57.2 & 61.5 \\
    Baseline (4 ep.) & 52.3 & 55.2 \\
    \midrule
    Recon. only (1 ep.) & 57.7 & 57.6 \\
    Recon. only (2 ep.) & 61.9 & 65.1 \\
    Recon. only (4 ep.) & 63.7 & 67.9 \\
    \midrule
    \rowcolor{Light!15}
    \textbf{\model~(Full, 1 ep.)} & \textbf{63.8} & \textbf{68.5} \\
    \bottomrule
    \end{tabular}}
\end{wraptable}

We compare our full model against a variant trained without the distillation losses ($\mathcal{L}_{\mathrm{token}}$ and $\mathcal{L}_{\mathrm{param}}$). As shown in Tab.~\ref{tab:abl_distill}, removing distillation~(see Recon. only 1 epoch) causes a substantial drop on both benchmarks. We empirically observe that without distillation, $\mathcal{L}_{\mathrm{recon}}$ remains large in scale even after 1 epoch, so a large fraction of the gradient signal is allocated to reducing $\mathcal{L}_{\mathrm{recon}}$, leaving the language modeling objective under-optimized. Consistent with this, the variant improves steadily with longer training ($1 \rightarrow2 \rightarrow 4$ epochs) and at 4 epochs nearly matches our full 1-epoch model. In contrast, the $\mathcal{L}_{\mathrm{LM}}$-only baseline shows only modest improvements with slightly longer iterations (1 $\rightarrow$2 epochs), and performance decreases after 4 epochs due to the model overfitting to the training data and losing general reasoning performance when only trained with $\mathcal{L}_{\mathrm{LM}}$. Together, these results indicate that the improved 3D reasoning comes primarily from jointly learning reconstruction and understanding, with distillation serving mainly to accelerate convergence and make training practical. 

\paragraph{\textbf{Isolating the source of the gains.}}
\begin{wraptable}[14]{r}{0.5\linewidth}
    \centering\footnotesize
    \vspace{-10pt}
    \caption{\textbf{Isolating the source of the gains.} \mbox{\emph{+ Teacher tokens}}: the teacher's query tokens are fed directly to the LLM as input. \emph{+ Distill. only}: summary tokens supervised by $\mathcal{L}_{\mathrm{distill}}$ but without $\mathcal{L}_{\mathrm{recon}}$.}
    \label{tab:abl_source}
    \resizebox{\linewidth}{!}{%
    \begin{tabular}{l cc}
    \toprule
    Variant & $\text{SQA3D}_{\text{test}}$ & SPAR-Bench \\
    \midrule
    Baseline & 56.5 & 60.9 \\
    \;+ Teacher tokens & 56.4 & 60.8 \\
    \;+ Distill. only & 56.6 & 60.5 \\
    \;+ Recon. (4 ep.) & 63.7 & 67.9 \\
    \rowcolor{Light!15}
    \textbf{Full (Ours, 1 ep.)} & \textbf{63.8} & \textbf{68.5} \\
    \bottomrule
    \end{tabular}}
\end{wraptable}
Having established that reconstruction, not longer training, drives the gains, we next ask whether the improvement truly comes from learning to reconstruct the scene~(mental reconstruction), or merely from access to the teacher's representation, the distillation signal, or the added token capacity. We disentangle these factors in Tab.~\ref{tab:abl_source}. First, feeding the teacher's query tokens directly to the LLM as additional input~(\emph{+ Teacher tokens}), without asking the model to reconstruct, performs on par with the baseline. Second, supervising the summary tokens with the distillation loss but \emph{without} the reconstruction loss~(\emph{+ Distill. only}) likewise fails to help. We attribute both negative results to a representational mismatch between the teacher's query tokens and the LLM's own embedding space: simply injecting or distilling these tokens does not teach the LLM how to interpret or attend to them, and without a dedicated mechanism to compensate for this misalignment or to explicitly force the language model to attend to the injected tokens, as in feature-fusion approaches such as VLM3R~\cite{fan2025vlm} the added tokens are largely ignored, leaving performance at the baseline level. In contrast, the reconstruction loss compels the Gaussian summary tokens to actively attend to the image tokens and extract the visual evidence required to estimate their 3D Gaussian parameters. This active aggregation reshapes the attention and representations \emph{within} the LLM such as sharpening cross-view correspondence and inducing object-level grouping~(\S\ref{subsec:analysis}), which we believe iss what benefits downstream reasoning. Together, these results indicate that the improvement arises primarily from jointly learning mental reconstruction.\vspace{-5pt}

\paragraph{\textbf{Number of Gaussian summary tokens.}}
\begin{wraptable}[8]{r}{0.4\linewidth}
    \centering\footnotesize
    \vspace{-13pt}
    \caption{
    \textbf{Number of Gaussian summary tokens $M$.}
    }
    \label{tab:abl_num_tokens}
    \begin{tabular}{l ll}
    \toprule
    $M$ & $\text{SQA3D}_{\text{test}}$ & SPAR-Bench \\
    \midrule
    1296 & 61.9 & 64.6 \\
    \rowcolor{Light!15}
    2592 (Ours) & \textbf{63.8} & \textbf{68.5} \\
    5184 & 60.3 & 58.4 \\
    \bottomrule
    \end{tabular}
\end{wraptable}

An important design choice of \model\ is that the Gaussian summary tokens act as a representational bottleneck, encouraging multi-view content corresponding to the same 3D region to be merged into shared tokens~(\S~\ref{subsubsec:gaussian_tokens}). To validate this design, we ablate the number of summary tokens $M$ while keeping all other components fixed. As reported in Tab.~\ref{tab:abl_num_tokens}, we compare introducing roughly 40, 80, 160 tokens per image, which makes the total number of summary tokens 1296, 2592, 5184, given 32 images. The evaluation shows that too few tokens~(1296) provide insufficient capacity, while too many tokens~(5184) weaken the bottleneck that drives object-level grouping, leading both extremes to underperform our chosen setting.
\vspace{-10pt}
\section{Conclusion}
We presented \model, a framework that equips a multi-image MLLM with the ability to perform an explicit \emph{mental reconstruction} of a 3D scene before answering. We introduce a compact set of learnable \emph{Gaussian summary tokens} that are appended to the image tokens and decoded into a compact 3DGS representation, jointly trained with the standard language modeling objective. Across challenging spatial reasoning benchmarks, \model\ significantly outperforms prior approaches that rely on explicit pixel-level geometry or geometry-foundation features, and our analyses further reveal that the reconstruction objective propagates beyond the Gaussian summary tokens, reshaping the image features into more spatially aligned and semantically structured representations. We believe that enabling machines to understand how the scene is structured before answering offers a promising step toward MLLMs that perceive 3D space in a more human-like manner.

\section*{Acknowledgements}
This work was supported by the Swiss AI Initiative through a grant from the Swiss National Supercomputing Centre (CSCS) under project ID ab036 on the Alps infrastructure; the Swiss National Science Foundation (Advanced Grant 216260: "Beyond Frozen Worlds: Capturing Functional 3D Digital Twins from the Real World"); the European Union’s Horizon Europe research and innovation programme under grant agreement No. 101214398 (ELLIOT); the Horizon Europe programme under grant agreement No. 101298421 (GRAIL); and the ETH AI Center through an ETH AI Center postdoc fellowship awarded to Sunghwan Hong.

% \clearpage

\small
\bibliographystyle{plainnat}
\bibliography{references}

%%%%%%%%%%%%%%%%%%%%%%%%%%%%%%%%%%%%%%%%%%%%%%%%%%%%%%%%%%%% Appendix
\clearpage
\appendix
This appendix provides additional implementation details~(\S~\ref{appendix:implementation_detail}), correspondence analysis~(\S~\ref{appendix:correspondence}), further analyses~(\S~\ref{appendix:analysis}), additional related works~(\S~\ref{appendix:rel_work}), and limitations~(\S~\ref{appendix:limitations}) that complement the main paper.
\section{Additional Implementation Details}
\label{appendix:implementation_detail}

\paragraph{Datasets and input.}
We use an input resolution of $384 \times 384$, matching the native resolution of LLaVA-Video-7B~\cite{zhang2024llavavideo}. For the ScanNet-based datasets~(SQA3D~\cite{ma2022sqa3d}, ScanQA~\cite{scanqa}, Scan2Cap~\cite{scan2cap}, ScanRefer~\cite{scanrefer}, and Multi3DRefer~\cite{zhang2023multi3drefer}), we uniformly sample 32 frames per scene, following prior work~\cite{llava3d,video3dllm,wang2025ross3d,fan2025vlm,hu2025g2vlm}. For SPAR~\cite{spar}, we use the 104K-sample multi-view subset released by VG-LLM~\cite{zheng2025learning}, in which each scene contains either 2, 3, or 32 views, and we feed all available views per scene to the model. We mix SPAR-7M into the ScanNet-based training set to mitigate the language-pattern overfitting tendency previously observed when training on ScanNet alone~\cite{ma2026real3dqa,zheng2025learning}.

\paragraph{Compact Gaussian teacher.}
We build our teacher on top of ZipSplat's~\cite{veicht2026zipsplat} open-source codebase, where we control the number of compact tokens to 2592 to make an adequate bottleneck structure as mentioned in \S~\ref{sec:ablation}. The teacher can be readily swapped with other token-level feed-forward 3DGS frameworks~\cite{an2025c3g,itkin2026globalsplat,ren2026tokengs}, and is kept frozen and used only at training time.

\paragraph{Training and architecture details.}
We train with the AdamW optimizer using a global batch size of 64 with 4 gradient accumulation steps. The peak learning rate after warmup is $1\mathrm{e}{-5}$ for the LLM and $2\mathrm{e}{-6}$ for the vision encoder, and we clip the gradient norm to $1.0$. For our full training objective, we set $\lambda_\text{recon}=3.0$ and $\lambda_\text{distill}=1.0$, with $\lambda_\text{MSE}=\lambda_\text{param}=\lambda_\text{token}=1.0$ and $\lambda_\text{LPIPS}=0.05$. The teacher-alignment projector $g(\cdot)$ in \S\ref{subsubsec:distillation} is a two-layer MLP with a GeLU activation~\cite{hendrycks2016gaussian} that maps the LLM's hidden dimension $3584 \rightarrow 1536$ to match the teacher's query dimension $d_T$. The Gaussian decoder head $\mathcal{H}(\cdot)$ is a three-layer MLP with SiLU activations~\cite{elfwing2018sigmoid} mapping $1536 \rightarrow 736$, which is the dimension required to specify the parameters of $G=32$ Gaussians per summary token. We use spherical harmonics of degree 2 for view-dependent color.

The 3D LMMs we compare against in Tab.~\ref{tab:3dqa}, Tab.~\ref{tab:caption_grounding}, and Tab.~\ref{tab:spar} differ in how they encode the 3D scene: Scene-LLM~\cite{scenellm} and LL3DA~\cite{ll3da} consume point-cloud features; LEO~\cite{leo}, ChatScene~\cite{huang2024chatscene}, and Grounded 3D-LLM~\cite{grounded-3dllm} use object-centric representations; LLaVA-3D~\cite{llava3d} lifts 2D features into a 3D voxel grid; and Video-3D-LLM~\cite{video3dllm}, GPT4Scene~\cite{gpt4scene}, and Ross3D~\cite{wang2025ross3d} treat multi-view images as video sequences.
\clearpage
\section{Additional Correspondence Analysis}
\label{appendix:correspondence}

\begin{figure}[t]
    \centering
    \includegraphics[width=1\linewidth]{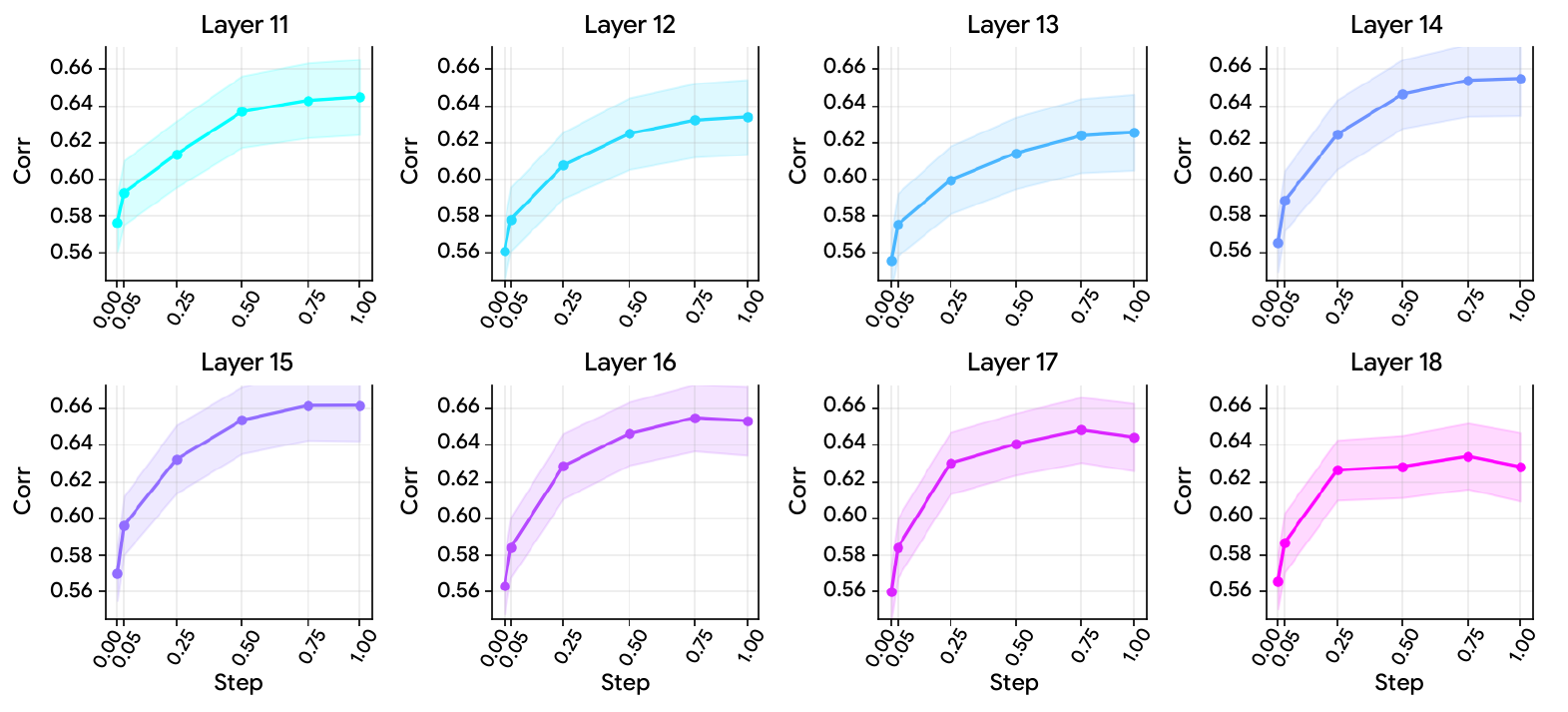}
    \vspace{-10 pt}
    \caption{\textbf{Layer-wise cross-view correspondence during training.}
    We report cross-view correspondence scores for image-token representations extracted from middle LLM layers $\ell =11 \sim 18$ on SQA3D~\cite{ma2022sqa3d} test samples. Scores increase consistently throughout training, with pronounced gains near the Gaussian decoding layer ($\ell=14$), indicating that joint reconstruction training improves the view-consistency of intermediate image features. Shaded regions indicate variance across SQA3D~\cite{ma2022sqa3d} test samples.}
    \label{fig:corr}
\end{figure}

Following 3DRS~\cite{huang20253drs}, we evaluate cross-view feature correspondence as a quantitative probe of how 3D-aware the image-token representations have become. We voxelize the 3D coordinates associated with image tokens with the given 3D point clouds for each scene of ScanNet~\cite{dai2017scannet}, construct cross-view pairs from tokens in different views that fall into the same voxel, and report the correspondence score as the average cosine similarity between the $\ell$-th layer hidden states of these pairs. A higher score indicates stronger feature consistency for image tokens describing the same 3D region across viewpoints. As shown in Fig.~\ref{fig:corr}, correspondence scores in the middle LLM layers, particularly the layers $\ell =11 \sim 18$ that surround the Gaussian decoding layer, improve substantially throughout training. This finding plays a dual role. As a corollary, it confirms that reconstruction supervision propagates 3D-aware structure beyond the summary tokens themselves into the image-token representations. We further attribute this propagation to our compact-bottleneck design ($M \ll KN'$): with far fewer summary tokens than image tokens, each summary token is forced to aggregate the same 3D region across multiple views, which in turn pressures the upstream image features to encode that region consistently across viewpoints. More importantly, the result shows that joint reconstruction \emph{delivers} the kind of cross-view correspondence that prior methods such as 3DRS supply through explicit supervision: improved correspondence emerges naturally from learning to reconstruct, without ever being directly optimized.
\section{Additional Analysis}
\label{appendix:analysis}
\subsection{Additional visualization of attention between images}
\begin{figure}[t]
    \centering
    \includegraphics[width=1\linewidth]{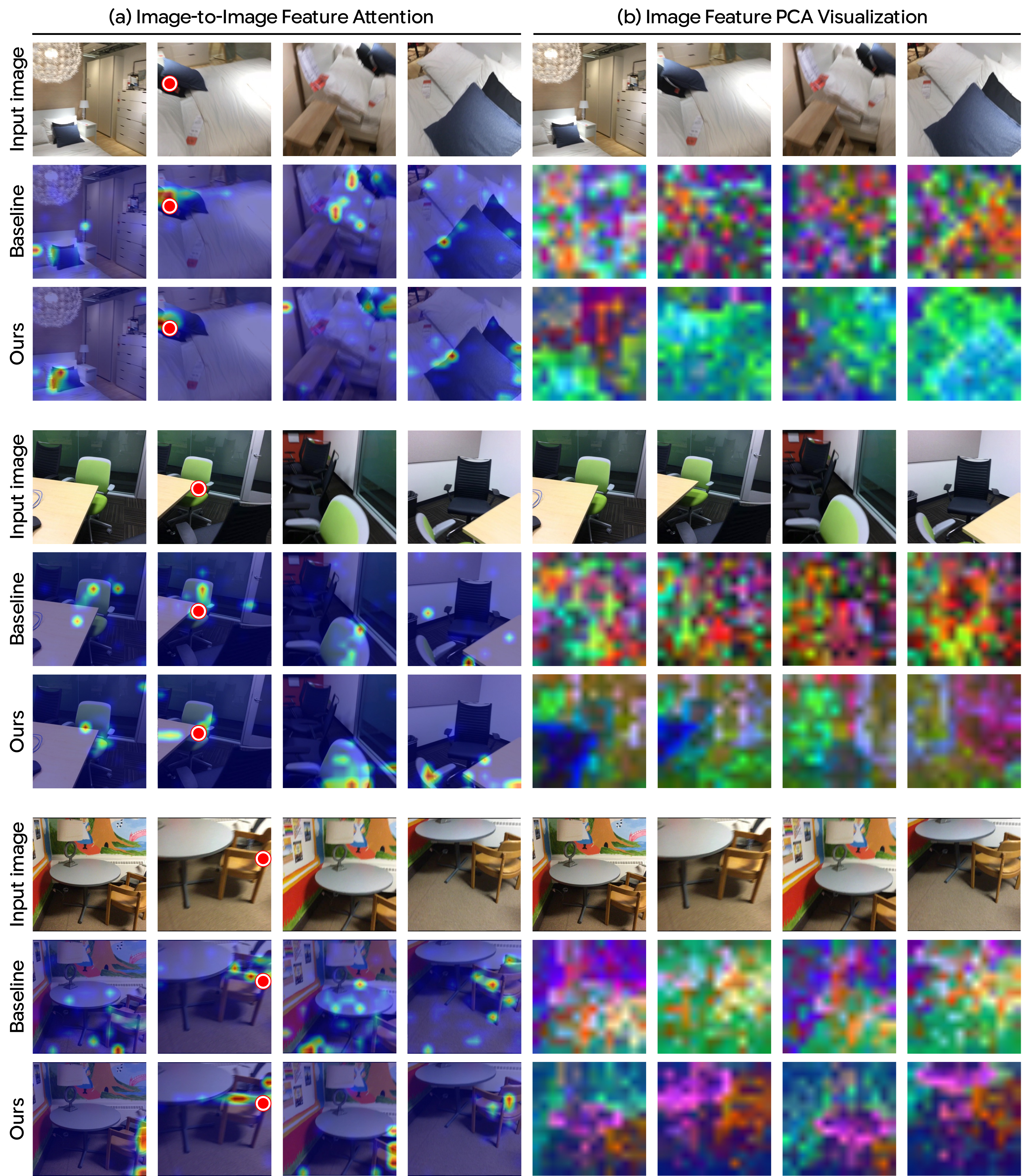}
    \vspace{-10 pt}
    \caption{\textbf{Visualization of attention maps between multi-view images and PCA visualizations of the image features.} \textbf{(a)} Image-to-image attention maps between a query point (\textcolor{red}{red} dot) and other input views. Our model exhibits substantially stronger cross-view correspondences than the baseline, indicating that the Gaussian summary tokens implicitly drive the image features to align across views. \textbf{(b)} PCA visualization of the LLM's image features (first three components shown as RGB). Our features are noticeably cleaner and more semantically structured, with corresponding objects (e.g., chairs, desks) encoded with consistent colors across views, reflecting the emergence of object-level abstractions.}
    \label{supfig:corr_pca}
    
\end{figure}

We further visualize the attention maps across multi-view images using the same protocol as Fig.~\ref{fig:corr_pca}. As illustrated in Fig.~\ref{supfig:corr_pca}, image tokens from our model reliably focus on matching regions of the same object across different views, demonstrating their robustness across diverse scenes.

\subsection{Additional visualization of clustered Gaussian}
\begin{figure}[t]
    \centering
    \includegraphics[width=1\linewidth]{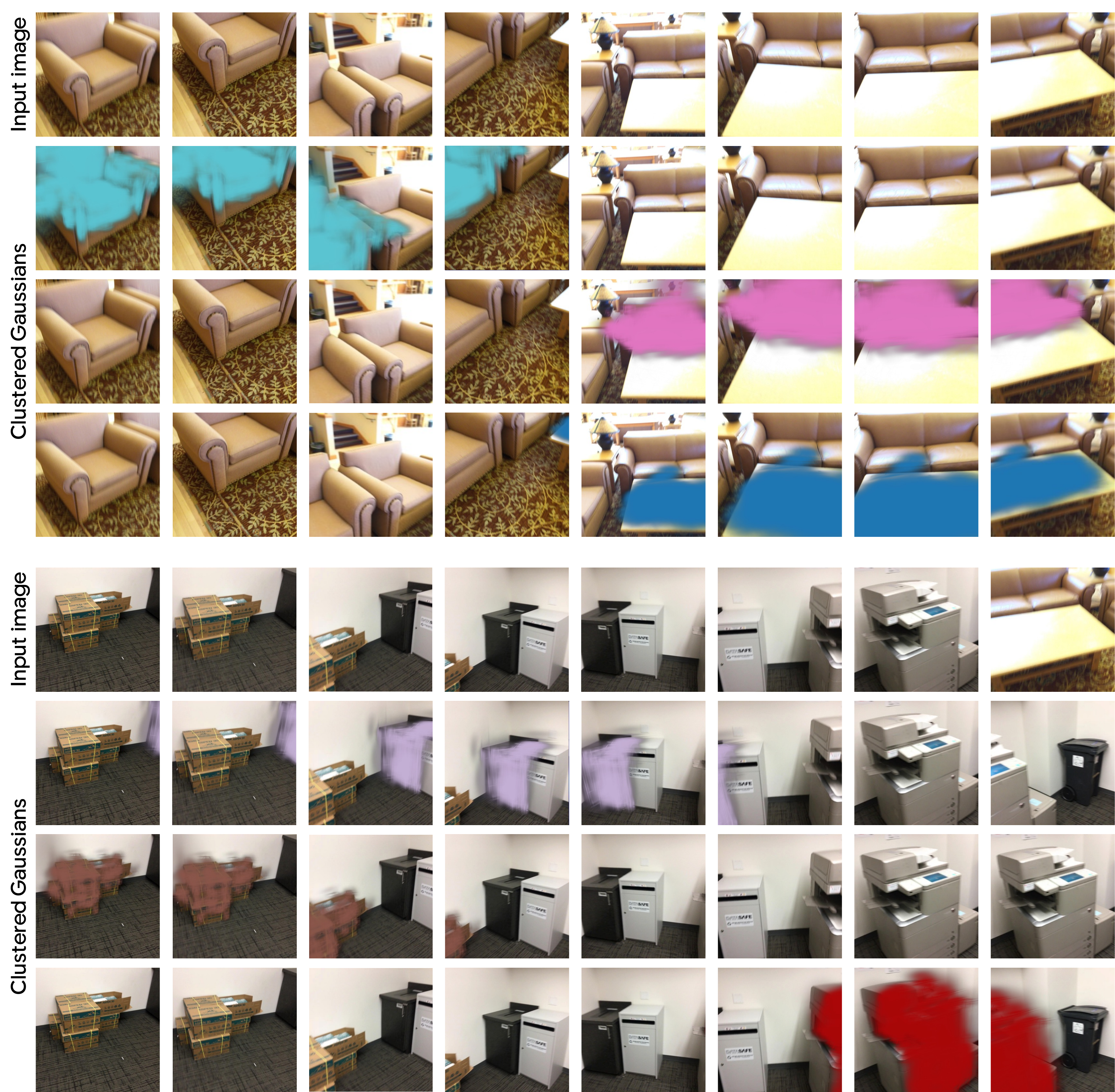}
    \vspace{-10 pt}
    \caption{\textbf{Visualization of clustered Gaussian from summary tokens.} We visualize the Gaussians clustered from a set of Gaussian summary tokens via K-means over the summary tokens. Similar summary tokens are mapped to the same object without any explicit clustering supervision.}
    \label{supfig:cluster_gaussian}
    
\end{figure}

We additionally visualize the attention maps between Gaussian summary tokens and multi-view images, following the same protocol as Fig.~\ref{fig:g2i}. As shown in Fig.~\ref{supfig:cluster_gaussian}, common objects are clustered into similar Gaussian summary tokens without any explicit supervision.

\subsection{Additional quantitative evaluation of semantic feature structuring}
\label{appendix:miou_probe}

The PCA visualizations in Fig.~\ref{fig:corr_pca} and the cross-view correspondence analysis in Appendix~\ref{appendix:correspondence} provide qualitative and geometric evidence that joint reconstruction training improves the LLM's image features. To obtain a more direct quantitative measure of whether the features become more \emph{semantically} structured, we conduct an unsupervised semantic segmentation probe following the standard evaluation methodology of~\cite{cuttano2026insid3, hamilton2022unsupervised}.
 
Specifically, on the ScanNet scenes of the SQA3D evaluation set~(which provide per-image 2D semantic labels), we extract the LLM's image features at layer $\ell$ for all 32 input views and apply K-means clustering \emph{jointly} over all views' tokens, with $k$ set to the number of ground-truth NYU40 classes present in the scene. Since the resulting clusters carry no labels, we follow the standard unsupervised segmentation protocol: each cluster is assigned to its maximally-overlapping ground-truth class via majority vote. After assignment, cluster labels are upsampled to the label resolution and we compute mIoU against the ground-truth semantic masks, aggregated across all views of each scene. As DINOv3~\cite{simeoni2025dinov3} features are well-known to be semantically structured, we additionally evaluate DINOv3 under the same protocol as an upper-bound reference.
 
As reported in Tab.~\ref{tab:miou_probe}, our model's image features achieve 35.43 mIoU, a substantial improvement over the baseline~(26.54) that closes much of the gap to DINOv3~(37.62). This result provides direct evidence that the feature space itself, not only its downstream usage, becomes more semantically structured through joint reconstruction training. Importantly, this structuring emerges as a byproduct of the reconstruction objective---the image features are never directly supervised with semantic labels---confirming that learning to reconstruct propagates object-level structure into the LLM's visual representations.

\begin{table}[t]
    \centering\small
    \caption{\textbf{Unsupervised semantic segmentation probe.} We cluster the LLM's image features via K-means and measure mIoU against ground-truth ScanNet labels. DINOv3~\cite{simeoni2025dinov3} is evaluated under the same protocol as an upper-bound reference.}
    \vspace{10pt}
    \label{tab:miou_probe}
    \begin{tabular}{l c}
    \toprule
    Method & mIoU \\
    \midrule
    Baseline (no summary tokens) & 26.54 \\
    \rowcolor{Light!15}
    \textbf{\model~(Ours)} & \textbf{35.43} \\
    \midrule
    DINOv3~\cite{simeoni2025dinov3} & 37.62 \\
    \bottomrule
    \end{tabular}
\end{table}

\subsection{Additional quantitative evaluation of reconstruction quality}
\label{appendix:recon_quality}
 
\begin{table}[t]
    \centering\small
    \caption{\textbf{Reconstruction quality on SQA3D test scenes.} We compare the rendering quality of our Gaussian summary tokens against the compact Gaussian teacher~\cite{veicht2026zipsplat} used during training. Higher is better for PSNR and SSIM; lower is better for LPIPS.}
    \vspace{10pt}
    \label{tab:recon_quality}
    \begin{tabular}{l ccc}
    \toprule
    Method & PSNR ($\uparrow$) & SSIM ($\uparrow$) & LPIPS ($\downarrow$) \\
    \midrule
    Gaussian teacher~\cite{veicht2026zipsplat} & 20.60 & 0.76 & 0.44 \\
    \rowcolor{Light!15}
    \model~(Ours) & 17.61 & 0.74 & 0.49 \\
    \bottomrule
    \end{tabular}
\end{table}
 
While our primary goal is to improve 3D reasoning rather than novel view synthesis, here we further analyze rendering quality on the SQA3D test scenes and compare against the compact Gaussian teacher~\cite{veicht2026zipsplat} in Tab.~\ref{tab:recon_quality}. As expected, our reconstruction quality is somewhat below the teacher's, since our Gaussians are decoded from the hidden states of an LLM that is jointly optimized for language modeling and reconstruction, trained with much fewer iterations, and without any special design to match the teacher's rendering fidelity. However, the primary goal of our design is not to achieve high-fidelity rendering: the reconstruction objective serves as an inductive bias that teaches the model how the scene is structured~(mental reconstruction), not as a novel view synthesis system~(\S~\ref{subsubsec:gaussian_tokens}).

\subsection{Additional analysis between Gaussian summary token and multi-view images}
\begin{figure}[t]
    \centering
    \includegraphics[width=1\linewidth]{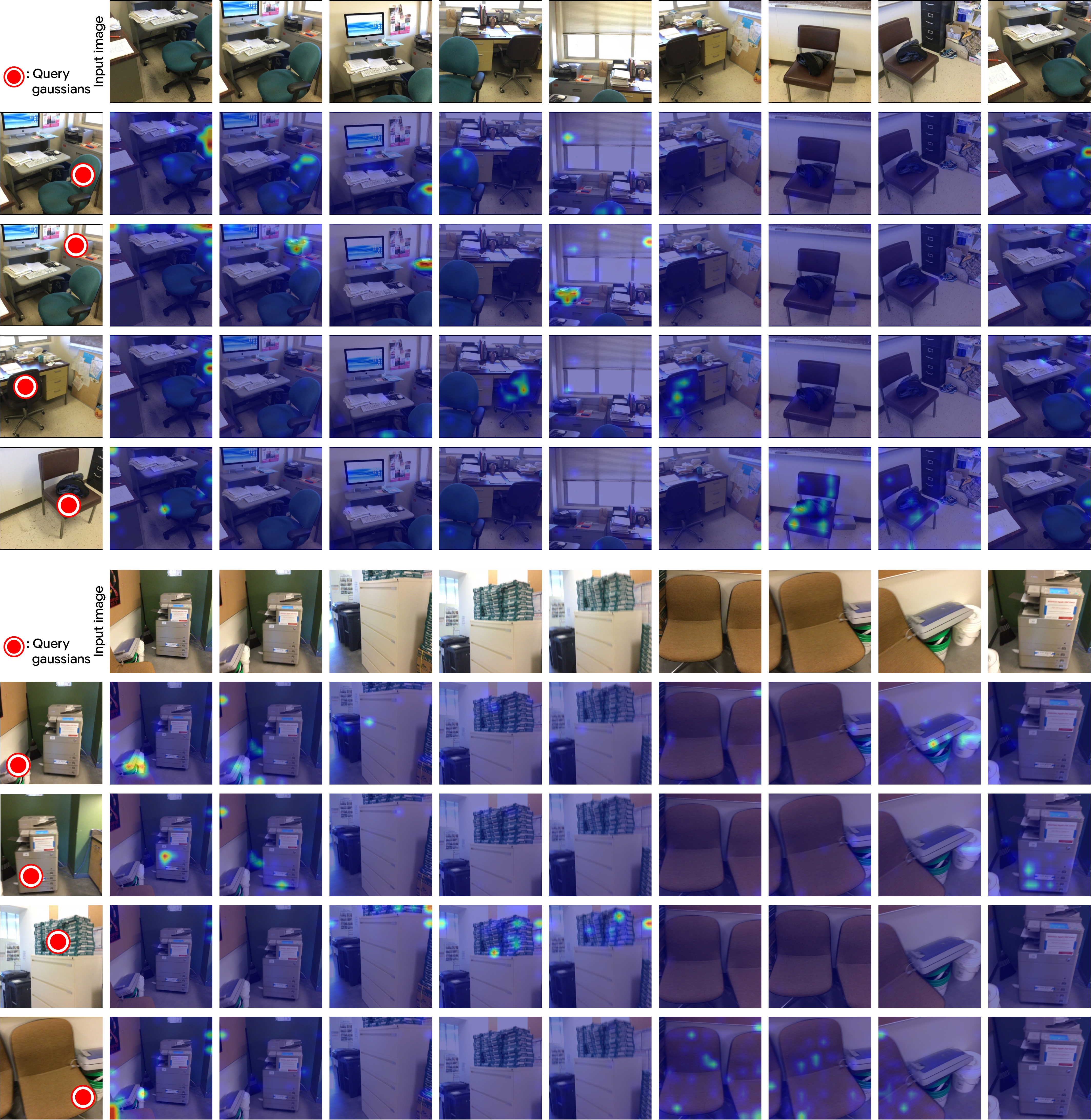}
    \vspace{-10 pt}
    \caption{\textbf{Visualization of attention maps between Gaussian summary tokens and images.} We visualize the attention score map between a Gaussian summary token (a \textcolor{red}{red} dot indicating the decoded Gaussian) and the input images. Each Gaussian token attends to corresponding regions of the same object across multiple views.}
    \label{supfig:g2i}
    
\end{figure}

To accurately reconstruct the scene, it is crucial for the Gaussian summary tokens to effectively aggregate information from multi-view images. We conduct additional analysis to examine how Gaussian summary tokens attend to multi-view image features when generating 3D Gaussians. As shown in Fig.~\ref{supfig:g2i}, each Gaussian summary token attends to corresponding regions of the same object across multiple views in the scene.
This behavior emerges because a limited number of Gaussian summary tokens are compelled to aggregate multiple occurrences of the same object into a single representation, effectively clustering common objects without any explicit supervision.
This emergent object-centric property is reminiscent of human perception, where the same object observed from multiple viewpoints is interpreted as a single entity to form a coherent 3D understanding of the scene.


\clearpage

\subsection{Computation cost analysis.}
\label{appendix:computation_cost}
\begin{wraptable}[13]{r}{0.45\linewidth}
    \centering\footnotesize
    \vspace{-12pt}
    \begin{tabular}{l c}
    \toprule
    Training setting & Peak VRAM~(GB) \\
    \midrule
    w/o $\mathcal{L}_{\mathrm{recon}} + \mathcal{L}_{\mathrm{distill}}$ & 40 \\
    \rowcolor{Light!15}
    \model~(full) & 44 \\
    \bottomrule
    \end{tabular}
    \caption{
    \textbf{Training-time computation cost.} Peak per-GPU VRAM measured during training under identical batch size and sequence length. The reconstruction and distillation losses introduce only a modest memory overhead while substantially improving 3D reasoning.
    }
    \label{tab:abl_compute}
\end{wraptable}
We further analyze the additional training-time cost incurred by our reconstruction and distillation objectives. As reported in Tab.~\ref{tab:abl_compute}, augmenting the standard language modeling loss with $\mathcal{L}_{\mathrm{recon}}$ and $\mathcal{L}_{\mathrm{distill}}$ raises the peak per-GPU VRAM from 40~GB to 44~GB, a 10\% increase under identical batch size and sequence length. This modest overhead is a direct consequence of our design choice to keep the number of Gaussian summary tokens compact~($M=2592$, $\S$~\ref{subsubsec:gaussian_tokens}). To make this concrete, pixel-aligned feed-forward 3DGS estimators~\cite{charatan2024pixelsplat, chen2024mvsplat, hong2024pf3plat, ye2024no} predict one Gaussian per input pixel, which for our setting of 32 images at $384\times384$ resolution would amount to roughly 4.7M Gaussians per scene. In contrast, \model\ decodes only $M\times G = 2592 \times 32 \approx 83\text{K}$ Gaussians, less than 1.8\% of the pixel-aligned count. This compactness keeps the rasterization, the teacher forward pass, and the gradient buffers comparatively lightweight, allowing the reconstruction objective to fit alongside MLLM finetuning at a manageable cost. The overhead is also confined to training, as the compact Gaussian teacher~\cite{veicht2026zipsplat} is discarded at inference time and the Gaussian decoder head $\mathcal{H}(\cdot)$ is only invoked when reconstructions are explicitly required. Given the substantial gains in 3D reasoning across all evaluated benchmarks~(Tab.~\ref{tab:3dqa},~\ref{tab:caption_grounding},~\ref{tab:spar}), we view this as a favorable trade-off, indicating that imagining the scene through a compact Gaussian abstraction is not only effective but also practically affordable.

\paragraph{\textbf{Inference time cost.}} 

We further report peak GPU memory and per-sample inference time on the SPAR evaluation set for three models: (a) the baseline multi-image MLLM taking only images, (b) VLM3R-7B~\cite{fan2025vlm}, which fuses image features with external CUT3R~\cite{wang2025continuous} features, and (c) our model.
 
As shown in Tab.~\ref{tab:inference_cost}, \model\ uses 20.43\,GB and 176.9\,ms per sample, compared to the baseline's 17.09\,GB and 112.1\,ms. The additional overhead compared to the baseline comes almost entirely from the longer input sequence~($M = 2592$ additional summary tokens), as the Gaussian decoder head is only invoked when the user explicitly requests a reconstruction and the distillation teacher is discarded after training. Importantly, approaches like VLM3R that fuse external 3D foundation model features must run a separate model~(CUT3R) on all input frames at inference before fusion, resulting in substantially higher memory~(25.29\,GB) and latency~(344\,ms). Our approach requires no external 3D model at inference, keeping the cost well below that of feature-fusion alternatives while still providing substantial reasoning gains.

\begin{table}[t]
    \centering\small
    \caption{\textbf{Inference cost comparison.} Peak GPU memory and per-sample inference time measured on the SPAR evaluation set. VLM3R-7B~\cite{fan2025vlm} additionally runs an external 3D foundation model~(CUT3R~\cite{wang2025continuous}) on all input frames before fusion.}
    \vspace{10pt}
    \label{tab:inference_cost}
    \begin{tabular}{l cc}
    \toprule
    Method & Peak Memory (GB) & Inference Time (ms) \\
    \midrule
    Baseline (7B) & 17.09 & 112.1 \\
    VLM3R-7B~\cite{fan2025vlm} & 25.29 & 344 \\
    \rowcolor{Light!15}
    \model~(7B) & 20.43 & 176.9 \\
    \bottomrule
    \end{tabular}
\end{table}

\clearpage

\subsection{Visualization of estimated Gaussians.}
\begin{figure}[t]
    \centering
    \includegraphics[width=1\linewidth]{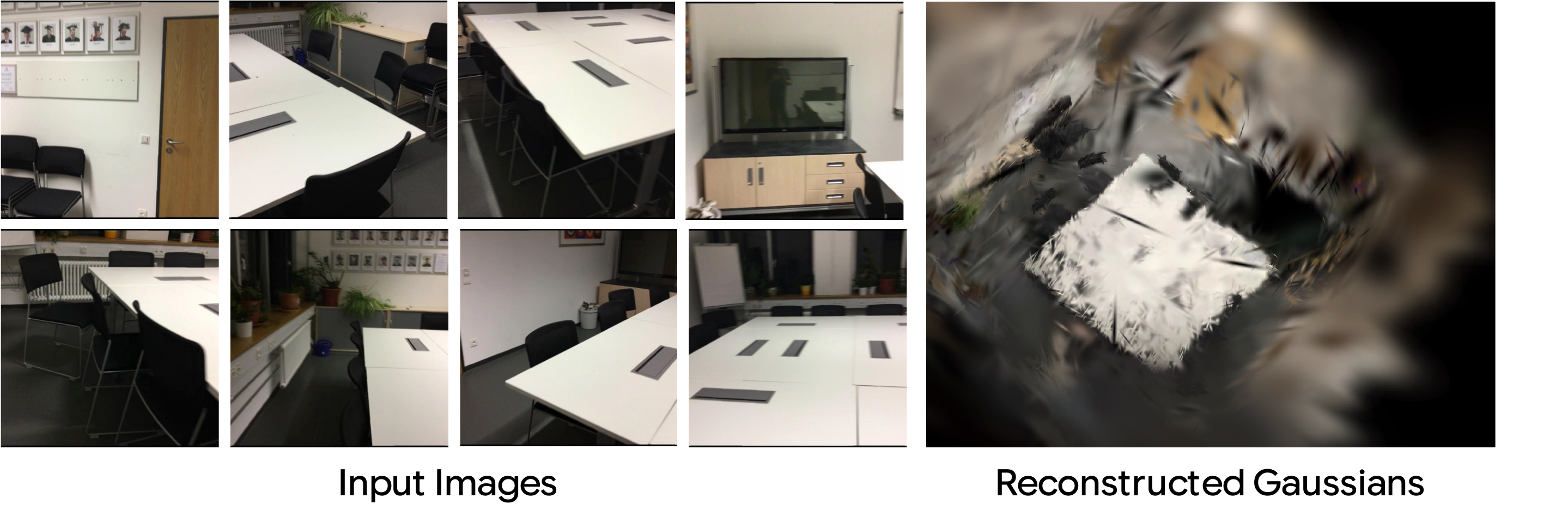}
    \vspace{-10 pt}
    \caption{\textbf{Visualization of reconstructed scene with estimated Gaussians.} 
\textbf{(Top)}: input multi-view images. \textbf{(Bottom)}: a view rendered from the 3D Gaussians 
decoded by our Gaussian summary tokens. Although the rendering is coarse due 
to the compact $M=2592$ token bottleneck, it preserves the overall scene layout 
and the placement of major objects, indicating that the tokens encode a 
coherent 3D abstraction of the scene.}
    \label{supfig:gauss_viz}
    
\end{figure}

While our main objective is improving 3D reasoning rather than achieving photorealistic novel view synthesis, we additionally visualize the estimated Gaussians in 3D to verify that the Gaussian summary tokens indeed acquire a coherent 3D understanding of the scene. As shown in Fig.~\ref{supfig:gauss_viz}, the rendered views recover the overall layout of the scene and the rough placement of major objects~(e.g.,~the central table surface and surrounding seating arrangement), confirming that even when decoded from only $M=2592$ summary tokens, the model is able to organize multi-view evidence into a spatially consistent 3D abstraction. We emphasize, however, that the rendering quality itself is not the primary concern of \model. Unlike feed-forward 3DGS estimators that are explicitly optimized for high-fidelity novel view synthesis~\cite{charatan2024pixelsplat, chen2024mvsplat, hong2024pf3plat, ye2024no, jiang2025anysplat}, our Gaussian summary tokens are trained jointly with the language modeling objective and are bottlenecked to a small set of tokens, which inherently trades off pixel-level fidelity for compact, object-level abstraction.

\clearpage
\section{Additional Related Works}
\label{appendix:rel_work}
Because our framework incorporates 3D Gaussian Splatting within an MLLM, several existing directions may appear superficially related to \model. We briefly discuss them here and clarify the key distinctions.

\paragraph{Feed-forward 3DGS as a feature lifter for MLLMs.}
SplatTalk~\cite{thai2025splattalk} also pairs a feed-forward 3DGS estimator with an MLLM, but uses 3DGS as an external \emph{feature lifter}: a frozen estimator~(FreeSplat~\cite{wang2024freesplat}) lifts pretrained 2D features into a 3D Gaussian field that is then fed to the LLM as visual tokens. The MLLM itself never learns to reconstruct the scene. In contrast, \model\ equips the MLLM with learnable Gaussian summary tokens that jointly learn to reconstruct the scene and support language modeling, with no external 3DGS module at inference.

\paragraph{Per-scene 3DGS-based scene understanding.}
LangSplat~\cite{qin2024langsplat}, FMGS~\cite{zuo2025fmgs}, and M3~\cite{zou20253d} distill 2D foundation features into 3D Gaussians optimized per scene, enabling open-vocabulary retrieval or segmentation. \model\ targets a different setting: a feed-forward generalist that answers natural-language questions about novel scenes in a single forward pass, without per-scene optimization.

\paragraph{Point-cloud-input MLLMs for 3D detection.}
SpatialLM~\cite{mao2025spatiallm} feeds 3D point clouds to an MLLM and outputs structured bounding-box coordinates and labels from a fixed vocabulary. \model\ instead targets free-form 3D understanding---spatial QA, situated reasoning, dense captioning, and grounding---through natural language interaction.

\paragraph{Latent geometric reasoning without explicit reconstruction.}
The most conceptually similar concurrent work is 3DThinker~\cite{chen2025think}, which also argues for ``thinking with 3D'' before answering spatial questions. 3DThinker aligns latent reasoning tokens with features from a frozen 3D foundation model~(VGGT~\cite{wang2025vggt}), placing it closer to methods that fuse geometry-foundation features into MLLMs~(e.g., VLM3R~\cite{fan2025vlm}, $\text{G}^2$VLM~\cite{hu2025g2vlm}). \model\ instead requires the Gaussian summary tokens to actually reconstruct the scene through differentiable rasterization, providing a stronger inductive bias toward 3D-aware internal representations~(\S\ref{subsec:analysis}). Empirically, \model\ outperforms 3DThinker-7B by 5.2 points on SPAR-Bench overall, with larger margins on the medium- and high-level reasoning splits.
\section{Limitations}
\label{appendix:limitations}
While \model\ demonstrates that abstract reconstruction is an effective inductive bias for 3D-aware MLLMs, training efficiency remains a limitation. As discussed in \S~\ref{subsubsec:distillation}, our framework can learn Gaussian summary tokens without a pretrained compact Gaussian teacher~\citep{veicht2026zipsplat}, but requires more training steps to achieve comparable downstream performance. The teacher therefore accelerates convergence rather than being a necessary component of our framework. Improving the efficiency of teacher-free training remains an important direction for future work.

Another limitation concerns the scope of our training scenes and the fixed representation budget. Our model is trained on indoor scenes, where a fixed number of Gaussian summary tokens and decoded Gaussians has not posed a major constraint in our experiments. However, larger and more complex outdoor scenes may require a greater representation capacity, and the current Gaussian budget could be insufficient to capture their spatial structure. Extending our framework to cover both indoor and outdoor environments, and dynamically adapting the number of Gaussian summary tokens to the scale and complexity of each scene, are promising directions for future work.

\clearpage

%%%%%%%%%%%%%%%%%%%%%%%%%%%%%%%%%%%%%%%%%%%%%%%%%%%%%%%%%%%% Appendix End

% \newpage
% \input{checklist.tex}

\end{document}